\documentclass[acmsmall]{acmart}
\usepackage{amsthm}
\newtheorem{definition}{Definition}[section]
\usepackage{tabularx}

\usepackage{graphicx}
\usepackage{booktabs}
\usepackage{appendix}
\usepackage{upgreek}
\usepackage{paralist}
\usepackage{verbatim}
\usepackage[caption=false,font=footnotesize,labelfont=sf,textfont=sf]{subfig}
\usepackage{hyperref}
\hypersetup{
colorlinks = true, 
linkcolor=black, 
citecolor=black, 
urlcolor=black 
}

\usepackage{float}
\usepackage{stfloats}
\usepackage{multirow}

\usepackage{amssymb}
\usepackage{amsmath}

\usepackage{xspace}
\usepackage{amsmath,amssymb,amsfonts}
\usepackage{algorithmic}
\usepackage{textcomp}
\usepackage{xcolor}
\usepackage{multicol}
\usepackage{wrapfig}
\usepackage{color, colortbl}
\definecolor{Gray}{gray}{0.9}
\usepackage[switch]{lineno}
\usepackage[ruled,vlined,linesnumbered]{algorithm2e}

\usepackage{makecell}

\usepackage[colorinlistoftodos,textsize=footnotesize]{todonotes}
\usepackage{float}
\usepackage[most]{tcolorbox}
\newtcbtheorem{Summary}{Conclusion}{
theorem name,
enhanced,
drop shadow={gray!20},
  coltitle=black,
  top=0.15in,
  left=0in,
  attach boxed title to top left=
  {xshift=1.5em,yshift=-\tcboxedtitleheight/2},
  boxed title style={size=small,colback=gray!60}
}{summary}

\AtBeginDocument{%
  \providecommand\BibTeX{{%
    \normalfont B\kern-0.5em{\scshape i\kern-0.25em b}\kern-0.8em\TeX}}}

\setcopyright{acmcopyright}
\copyrightyear{2023}
\acmYear{2023}
\acmDOI{10.1145/3582571}

\acmJournal{TOSEM}
\acmVolume{32}
\acmNumber{5}
\acmArticle{113}
\acmMonth{7}
\newcommand{\method}{\textsf{LATTICE}\xspace}
\newcommand{\pdm}{\textsf{LATTICE-PDM}\xspace}
\newcommand{\cem}{\textsf{LATTICE-CEM}\xspace}
\newcommand{\hdm}{\textsf{LATTICE-HDM}\xspace}
\newcommand{\cemhdm}{\textsf{LATTICE-CEM-HDM}\xspace}
\newcommand{\pdmcemhdm}{\textsf{LATTICE-CL}\xspace}
\newcommand{\latticedtm}{\textsf{LATTICE-DTM}\xspace}

\begin{document}
\title[Digital Twin-based Anomaly Detection with Curriculum Learning in Cyber-physical Systems]{Digital Twin-based Anomaly Detection with Curriculum Learning in Cyber-physical Systems} 
\author{Qinghua Xu}
\orcid{0000-0001-8104-1645}
\affiliation{%
\institution{Simula Research Laboratory}
\city{Oslo}
\country{Norway}}
\affiliation{%
\institution{University of Oslo}
\city{Oslo}
\country{Norway}}
\email{qinghua@simula.no}
\author{Shaukat Ali}
\orcid{0000-0002-9979-3519}
\affiliation{%
\institution{Simula Research Laboratory}
\city{Oslo}
\country{Norway}}
\email{shaukat@simula.no}
\author{Tao Yue}
\authornote{Corresponding author}
\orcid{0000-0003-3262-5577}
\affiliation{%
\institution{Simula Research Laboratory}
\city{Oslo}
\country{Norway}}
\email{tao@simula.no}


\begin{abstract}
Anomaly detection is critical to ensure the security of cyber-physical systems (CPS). However, due to the increasing complexity of attacks and CPS themselves, anomaly detection in CPS is becoming more and more challenging. In our previous work, we proposed a digital twin-based anomaly detection method, called ATTAIN, which takes advantage of both historical and real-time data of CPS. However, such data vary significantly in terms of difficulty. Therefore, similar to human learning processes, deep learning models (e.g., ATTAIN) can benefit from an easy-to-difficult curriculum. To this end, in this paper, we present a novel approach, named digitaL twin-based Anomaly deTecTion wIth Curriculum lEarning (\method), which extends ATTAIN by introducing curriculum learning to optimize its learning paradigm. \method attributes each sample with a difficulty score, before being fed into a training scheduler. The training scheduler samples batches of training data based on these difficulty scores such that learning from easy to difficult data can be performed. To evaluate LATTICE, we use five publicly available datasets collected from five real-world CPS testbeds. We compare \method with ATTAIN and two other state-of-the-art anomaly detectors. Evaluation results show that \method outperforms the three baselines and ATTAIN by 0.906\%-2.367\% in terms of the F1 score.  \textcolor{black}{\method also, on average, reduces the training time of ATTAIN by 4.2\% on the five datasets and is on par with the baselines in terms of detection delay time.}
\end{abstract}

\ccsdesc[500]{Software and its engineering~Maintaining software}
\ccsdesc[500]{Security and privacy~Intrusion detection systems}
\ccsdesc[500]{Computer systems organization~Embedded systems}
\ccsdesc[500]{Computer systems organization~Sensors and actuators}
\ccsdesc[500]{Computer systems organization~Embedded and cyber-physical systems}
\ccsdesc[500]{Computing methodologies~Neural networks}
\ccsdesc[500]{Computing methodologies~Supervised learning by classification}
\ccsdesc[500]{Computing methodologies~Online learning settings}
\keywords{cyber-physical system, digital twin, curriculum learning, deep learning, anomaly detection}

\maketitle
\section{Introduction} 
\textcolor{black}{Cyber-physical systems (CPS) have been deployed in many applications~\cite{Luo2020}}. These systems are also becoming more complex, heterogeneous, and integrated, i.e., consisting of multiple systems to provide rich functionalities, which expose CPS to broader threats. Such threats can cause significant damage to CPS. Therefore, anomaly detection, which can potentially detect these threats in advance, is a crucial task in the domain of CPS security.
 
Multiple anomaly detection mechanisms for CPS have been proposed~\cite{Eckhart2019,Banerjee2011,Goh2017}. These traditional mechanisms include invariant checking, physical attestation, and fingerprinting~\cite{Eckhart2019}. Recently, deep learning-based anomaly detectors are intensively studied due to their success in various domains~\cite{Liu2019}, such as image classification and natural language processing. Typically, anomaly detectors are built as deep learning models and trained on static data collected from a CPS. Despite the success of these neural network-based anomaly detectors, most of them still suffer from two significant challenges. First, most of these models are trained on static log data and cannot keep learning during the operation of CPS. Consequently, these models perform the best when dealing with known attacks but suffer from low accuracy when facing novel attacks. Second, neural network-based models rely on a large amount of labeled data for training, which is known to be expensive in CPS ~\cite{Goh2017}. Comparably, a model that can take advantage of unlabeled data is, therefore, much more desirable and appreciated in practice. 

To tackle the above-mentioned challenges, in our previous work, we proposed ATTAIN~\cite{xudigital}, which takes advantage of unlabeled data and continuously learns during the operation of a CPS, i.e., at runtime. ATTAIN achieved new state-of-the-art results on benchmark datasets by taking advantage of both historical and real-time data introduced in chronological order. \textcolor{black}{Especially, training on live data keeps improving the predictive performance and efficiency of the digital twin, allowing it to make more accurate and earlier predictions of anomalies.}

Recently, the latest research on Curriculum Learning (CL) demonstrates that re-organizing the training data as a curriculum can boost the performance of deep learning models~\cite{Cirik2017,choi2019pseudo,Soviany2021,koenecke2019curriculum,kocmi2017curriculum}. CL is inspired by a key observation from human learning processes: humans acquire knowledge via performing a series of tasks, usually from easy to difficult tasks. In essence, human learning is typically organized as curricula. Similarly, machine learning algorithms can also benefit from CL. According to extensive research in this area, CL is proven to be effective in various machine learning domains, such as computer vision and natural language processing~\cite{Xu2020,Selfridge,Jiang2018,choi2019pseudo,Carpuat2017}. However, few works focus on exploring CL with time-series data from CPS, which is inherently chronologically ordered~\cite{koenecke2019curriculum}. Most existing works utilize sequential deep learning models to process these time-series data directly, without making any changes to the order of training data~\cite{Li2019}.

In this paper, we extend ATTAIN by employing CL and form the following new contributions:
\textcolor{black}{
\begin{itemize}
    \item We designed \textbf{a generic framework} that combines CL with ATTAIN. Next, we developed novel difficulty measurers and adapted CL to CPS time series data. In particular, we proposed two types of difficulty measurers: predefined and automatic difficulty measurers, which focus on the property of samples and are based on the context of the samples, respectively.
    \item We conducted \textbf{extensive empirical study} with five case studies, among which, as compared to ATTAIN, two case studies are newly added. 
    These two case studies further demonstrate the generalizability of \method over diverse datasets.
    \item We performed \textbf{extensive analyses with statistical tests} to evaluate \method. We first evaluate \method with coarse-grained and fine-grained effectiveness metrics, followed by investigating the effectiveness of CL and its components and the efficiency of \method. 
    Moreover, we discussed \textbf{the plausible reasons for the improvement}, including taking advantage of predefined difficulty measurer, automatic difficulty measurer, and CL's optimization strategy.
\end{itemize}
}

In total, we evaluate the cost-effectiveness of \method on five datasets collected from real-world critical infrastructure testbeds: Secure Water Treatment (SWaT)~\cite{Mathur2016}--- a multi-stage water purification plant, Water Distribution (WADI)~\cite{Ahmed2017}--- a consumer distribution network, Battle of Attack Detection Algorithms (BATADAL) ~\cite{taormina18battle}--- a dataset designed for an attack detection contest, PHM challenge 2015 dataset~\cite{phm} and Gas Pipeline Dataset~\cite{Morris2015}. We demonstrate that \method improves the performance of the state-of-the-art anomaly detection methods by 0.906\%, 2.363\%, 2.712\%, 2.008\%, 2.367\%, respectively, in terms of the F1 score. \textcolor{black}{We also evaluate the time efficiency of \method with Unit Training Time (UTT) and Detection Delay Time (DDT). Experiment results show that LATTICE improves UTT by 4.2\% and DDT by 0.2\% on average.}

The remaining part of this paper is organized as follows: Section ~\ref{sec:background} introduces the background of CL, digital twin, and Generative Adversarial Networks (GAN). We present related works in Section~\ref{sec:relatedwork} a running example in  Section~\ref{sec:runningexample}. Section~\ref{sec:approach} introduces LATTICE in detail. Section~\ref{sec:experimentDesign} shows our empirical evaluation, followed by Section~\ref{sec:Results} where we present the experimental results and analysis. Section~\ref{sec:threats} shows possible threats to validity in our experiments.
Section~\ref{sec:disscuss} provides the overall discussions. Finally, Section~\ref{sec:conclusion} summarizes the paper and proposes potential future work. A replication package of the experiments is provided here \footnote{https://github.com/xuqinghua-China/tosem/tree/master}.

\section{Background} \label{sec:background}
In this paper, we combine two main techniques: CL and digital twin. We first present a general CL framework in Section \ref{subsec:bg_cl},  while in Section \ref{subsec:bg_dt} we present the concept of digital twin and its components. Inside digital twin, we use GAN as the backbone for our model. In Section \ref{subsec:bg_gan}, we introduce the basic structure and key functions of GAN.

\subsection{Curriculum learning}\label{subsec:bg_cl}
CL was first introduced into machine learning models by Selfridge et al.~\cite{Selfridge}. In that paper, CL was used to address the cart pole controlling task, which is a classic problem in robotics. This work later inspires many researchers to explore CL for various tasks such as grammar learning~\cite{Krueger2009,Rohde1999,Krueger2009} and language learning ~\cite{bengiocl}. Bengio et el~\cite{bengiocl} first proposed the concept of CL in the domain of language learning. Below is the definition of \textit{Curriculum} and CL.
\begin{definition}\label{def:curriculum}
A curriculum is a sequence of training criteria over T training steps: $C=<Q_1,...,Q_t,...,Q_T>$. Each criterion $Q_t$ is a reweighting of the target training distribution $P(z)$:
\begin{equation}
    Q_t(z)\propto W_t(z)P(z)\qquad \forall{example}\,z \in training\,set\,D
\end{equation}
under the condition that the entropy of the distribution and weights, for any example, increase. Also, in the final steps, all examples are uniformly sampled.
    
\end{definition}
With the definition of \textit{Curriculum}, Bengio et el.~\cite{bengiocl} defined CL as follows. 
\begin{definition}\label{def:cl}
Curriculum learning is a training strategy that trains a machine learning model with a curriculum defined above. 
\end{definition}
 Most of the studies on CL follow this concept~\cite{bengiocl,Platanios2019,Wei2017,kocmi2017curriculum}. Wang et. al.~\cite{Wang} summarized these works and proposed a general framework of CL as "Difficulty Measurer + Training Scheduler". The difficulty measurer decides the relative "easiness" of each data sample, while the training scheduler decides the sequence of data subsets throughout the training process, based on the judgment of the difficulty measurer. In this paper, we follow this framework and define our own difficulty measurer and training scheduler.
 
\subsection{Digital Twin}\label{subsec:bg_dt}
\begin{figure}[!htb]
\centering
\includegraphics[width=0.4\columnwidth]{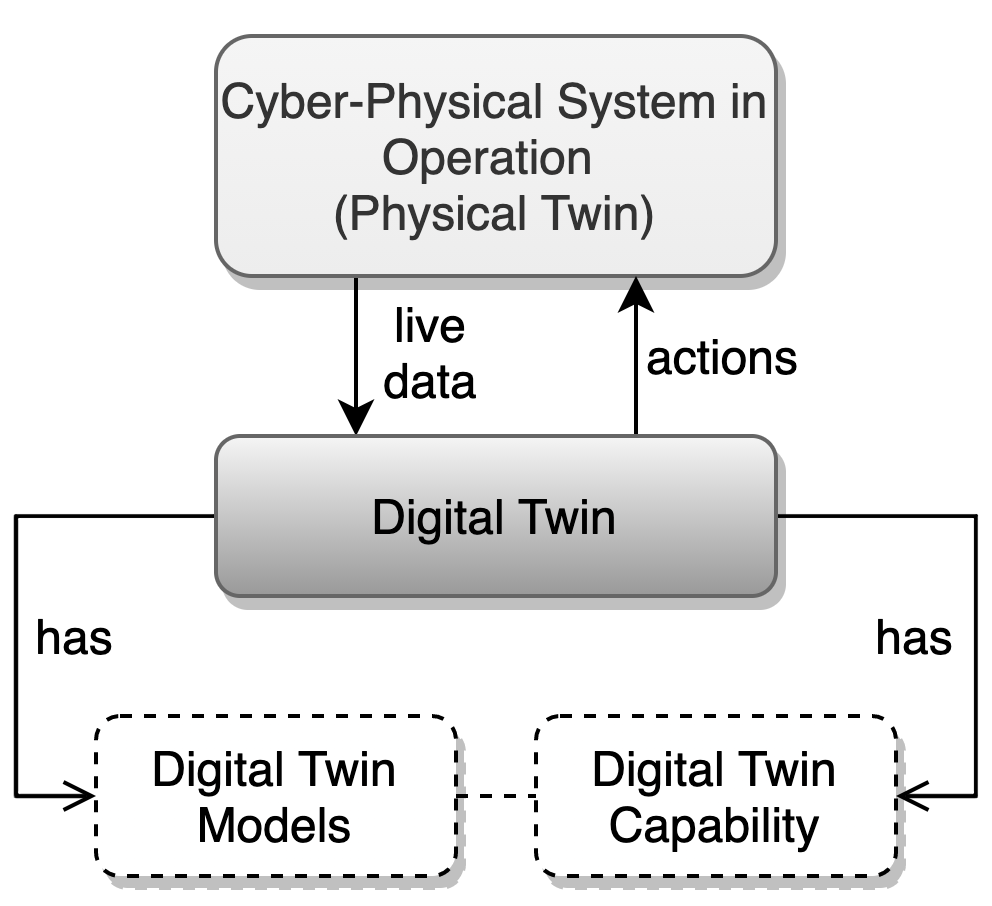}
\caption{Digital Twin for Cyber-Physical System}
\label{fig:dt}
\end{figure}

El Saddik~\cite{Eckhart2018} defined the concept of \textit{Digital Twin} as \textit{``a digital replica of a living or non-living physical entity''}. In this paper, we focus on building a digital twin of a CPS. Fig~\ref{fig:dt} shows a high-level view of a digital twin for a CPS in operation, which is referred to as a physical twin - a commonly used term in the literature~\cite{Eckhart2019,yue2022towards,yue2021understanding}. The data from the physical twin is continuously fed to the digital twin. Depending on the CPS, such data may come from its environment, communication medium, and the CPS itself. A digital twin may also be able to take action on the physical twin. Examples of such action include preventing unsafe situations or alerting a CPS operator about abnormal behaviors.

\textcolor{black}{As shown in Fig~\ref{fig:dt}, a digital twin consists of two parts: the  digital twin model (DTM) and the digital twin capability (DTC). The \textit{DTM} refers to the digital representation of the CPS. This representation could be in the form of heterogeneous models corresponding to different components such as software, hardware, and communication. We treat these components as black boxes and build the DTM with a data-driven approach. These models can be represented as state machines, machine learning models, etc. The \textit{DTC} refers to the functionality of a digital twin. Depending on the context, a digital twin can provide various functionalities, such as predictions of non-functional properties, uncertainty detection, and failure prevention. In our context, we focus on the anomaly detection capability that detects abnormal patterns exhibited by sensors and actuators and usually caused by external attacks. Moreover, the \textit{DTC} interacts with the \textit{DTM}, e.g., to perform simulations on it. Similarly, the \textit{DTM} interacts with the \textit{DTC}, e.g., to get feedback about whether or not the system is under attack. This bidirectional relationship is shown as a dashed line between the \textit{DTC} and the \textit{DTM} in Fig~\ref{fig:dt}. 
}

\textcolor{black}{
Traditionally, digital twins are constructed statically with historical data only \cite{Eckhart2018}. However, in some instances, we may have CPS design models that could be used and improved incrementally during the operation of the digital twin based on live data, thereby representing the most up-to-date state of the underlying CPS. \textcolor{black}{In particular, as discussed in our previous work ATTAIN~\cite{xudigital}, such live data improves both predictive performance and the efficiency of the digital twin. Better predictive performance indicates that the digital twin can make more accurate anomaly predictions, while better efficiency indicates that it takes less time for a digital twin to detect an anomaly. 
}}

\textcolor{black}{In this paper, we build the DTM as a timed automaton machine, and the backbone of the DTC is GAN. The DTM provides ground truth labels to improve the anomaly detection capability of \method. As for training, we train such models initially with historical data, followed by their continuous improvement with live data.}

\subsection{Generative Adversarial Networks}\label{subsec:bg_gan}
 As mentioned in Section \ref{subsec:bg_dt}, we utilize GAN as the backbone of DTC. GAN was invented by Ian Goodfellow in ~\cite{Goodfellow2014} and has been successful in performing many data generation tasks~\cite{Jabbar2020} \textcolor{black}{e.g., image generation~\cite{bao2017cvae} and text generation~\cite{zhang2017adversarial}. These successful applications demonstrate GAN's generalizability and inspire us to adapt it to the CPS security domain. } 
 
\begin{figure}[!htb]
\centering
\includegraphics[width=0.7\columnwidth]{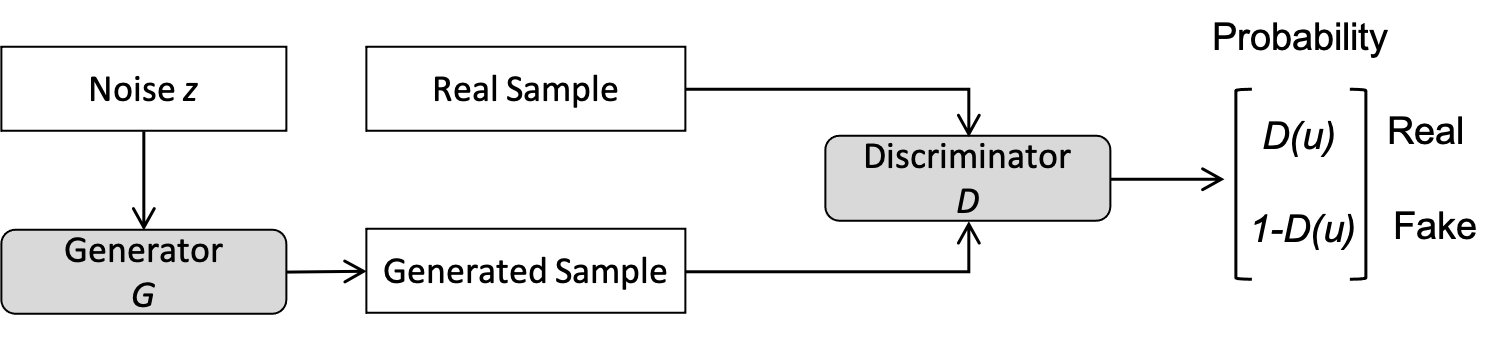}
\caption{Generative Adversarial Network Structure}
\label{fig:gan}
\end{figure}

 A typical GAN has independent models: generator $(G)$ and discriminator $(D)$ (see the GAN structure in Figure \ref{fig:gan}). Considering an input noise variable $z$, the goal of $G$ is to generate a new adversarial sample $G(z)$ that comes from the same distribution as of $u$. On the other hand, the discriminator model $(D)$ returns the probability $D(u)$ assessing whether the given sample $u$ is from the real data set or generated by $G$. The ultimate goal of $G$ is to maximize the probability that $D$ would mistakenly predict generated data as a real one, and the goal of $D$ is the opposite. Thus, $G$ learns to generate more realistic adversarial samples while $D$ continuously improves its capability of distinguishing them from real samples.
\section{Related Work} \label{sec:relatedwork}
In Section \ref{subsec:CPS AD}, we present related works about CPS anomaly detection.  In Section \ref{subsec:CL}, we present the research progress of CL and how to incorporate CL and digital twin.

\subsection{CPS Anomaly Detection}\label{subsec:CPS AD}
\subsubsection{Traditional Anomaly detection}
Anomaly detection is a popular research topic in the domain of CPS security. \textcolor{black}{Many traditional anomaly detection methods attempt to learn normal states of CPS with predefined rules(e.g., frequency limit), state estimation (e.g., the Kalman filter), and statistical models (e.g., Gaussian model, histogram-based model) ~\cite{Luo2020}. However, a disadvantage of these methods is the requirement for domain knowledge or data distribution of normal data. On the other hand, machine learning approaches have proved successful in various domains by leveraging data instead of domain knowledge to improve performance. Following this research line, our method adapts machine learning techniques to the CPS domain.}  


\subsubsection{Deep Learning-based Anomaly Detection}
Deep learning-based anomaly detection methods have been proposed to identify anomalies in CPS by exploring various neural network architectures to detect attacks in different CPS domains ~\cite{Luo2020}. Jonathan et al. ~\cite{Goh2017} introduced Long Short Term Memory Networks (LSTM) to capture temporal characteristics of time series data. LSTM was used as a predictor to model the normal behavior of the CPS, and subsequently, CUSUM was used to identify abnormal behaviors. Several approaches (e.g., ~\cite{Canizo2019,Kravchik2018,Yao2017}) adopt a convolutional layer as the first layer of a neural network to obtain correlations of multiple sensors in a sliding time window. Further, extracted features are fed to subsequent layers to generate output scores. Canizo et al. ~\cite{Canizo2019} and Wu et al.~\cite{Wu2018} used Recurrent Neural Network (RNN) to take the output of the Convolutional Neural Network (CNN) layer and form the prediction layer. Moreover, both methods used datasets from real industrial plants and applied precision, recall, F1, and ROC as the evaluation metrics. Such methods prove the effectiveness of deep learning techniques in various domains. We follow this research line and use LSTM to capture temporal characteristics of CPS data while utilizing Graph Convolution Network (GCN) instead of CNN to capture their spatial characteristics better. However, deep learning methods work best when trained on large amounts of labeled data, which can be costly to acquire in the CPS domain. Therefore, we take advantage of a digital twin for data augmentation.

\subsubsection{Digital Twin-based Anomaly Detection}
A few papers have proposed to use digital twins for anomaly detection. Eckhart \& Ekelhart ~\cite{Eckhart2018} built a knowledge-based intrusion detection system with digital twins, which is based on the assumption that a CPS would exhibit certain unusual behavior patterns during an attack. To this end, they proposed rules that the system must adhere to under normal conditions. Given these rules, they built a simple digital twin, which continuously checks rule violations at runtime. This method achieved a low false-positive rate in anomaly detection tasks. However, these rules need to be predefined and do not evolve or get updated when additional real-time data is available from operating systems as \method does. The reason is that their digital twin is simply a static representation of a real system, implying that the digital twin does not evolve/learn when new data are available.

Later on, Eckhart \& Ekelhart~\cite{Eckhart2018b} further improved their digital twin by introducing a passive state replication approach to simulate real systems with real-time data. To demonstrate the viability of the proposed digital twin, Eckhart \& Ekelhart (2018b) implemented a behavior-specification-based intrusion detection system. They evaluated the effectiveness by launching a man-in-the-middle, and an insider attack against a real CPS ~\cite{Eckhart2019}. Evaluation results show that the proposed anomaly detector yields a low false-positive rate while being capable of detecting unknown attacks. In their work, the specification of a CPS is used to automatically build the digital twin model, which is a Finite State Machine (FSM). FSM does not consider time constraints for transitions as Timed Automaton does, making it difficult to replicate a CPS with delayed transitions. Also, their intrusion detector depends on the digital twin model to simulate the correct behavior of a CPS. When a mismatch between a state of the digital twin model and the corresponding state of the real operating CPS occurs, the CPS is considered to be under attack. \textcolor{black}{However, mismatches are determined by predefined rules, which cannot detect complicated attack patterns, such as attacks with delayed effects and attacks targeting multiple access points simultaneously. \method, however, mitigates this challenge by using Timed Automaton as the digital twin model and GCN as the GAN generator to capture temporal characteristics.}

\subsection{Curriculum Learning}\label{subsec:CL}

Based on the framework proposed in ~\cite{Wang}, existing CL methods can be divided into predefined CL and automatic CL.
 
In predefined CL, the difficulty measurer and training scheduler are completely designed based on prior human knowledge without any data-driven models or algorithms involved. The difficulty measurers of a predefined CL need to be manually designed based on characteristics of specific data. Since CL was originally designed for computer vision and natural language processing tasks, most predefined difficulty measurers are related to images or text data, such as complexity~\cite{Platanios2019,Wei2017}, diversity~\cite{bengiocl,kocmi2017curriculum}, and noise estimation~\cite{choi2019pseudo,Chen2015}. Unlike predefined difficulty measurers, predefined training schedulers are usually task/data agnostic. In general, predefined training schedulers can be divided into two types: discrete and continuous training schedulers. The most popular discrete scheduler is called Baby Step~\cite{bengiocl}. The Baby 
Step algorithm first divides data into multiple buckets based on the difficulty score of each sample. The training process starts with the easiest bucket and slowly includes harder buckets after several training epochs. Other discrete schedulers, including One-Pass~\cite{bengiocl} and modified Baby Step~\cite{kocmi2017curriculum}, are also widely adopted due to their demonstrated simplicity and effectiveness.

\textcolor{black}{Despite its success, predefined CL is limited to domain knowledge and prior information requirements, which might not be accessible in certain application contexts. Also, predefined CL stays fixed during training, is incapable of adapting to potential novel scenarios during operation.} Therefore, automatic CL has attracted much attention recently. Self-paced Learning (SPL)~\cite{Tullis2011,Kumar} is a primary branch of automatic CL. The intuition behind SPL originates from human learning, where a student teaches her/himself, and controls the content, method, time, and length of the study. For instance, Kumar et al.~\cite{Kumar} proposed an SPL approach that trains the model at each iteration with the easiest subset based on the model's current performance, i.e., the examples with the lowest training losses. Another type of automatic CL is Transfer Teacher (TT)~\cite{Xu2020,Hacohen2019,Weinshall2018}. Unlike SPL, which depends on the student to teach itself, TT takes another pre-trained model as the teacher and transfers "knowledge" to the student model. SPL has a risk of uncertainty at the beginning of training if the student model is not mature enough to teach itself, while TT reduces this risk by inviting a mature teacher to help the student model for assessment. We follow this research line and use DTM as the teacher model and DTC as the student model, allowing LATTICE to adapt its curricula automatically and continuously.

\section{Running Example from the SWaT Testbed} \label{sec:runningexample}
To better illustrate our method, we present a running example in Table \ref{tab:running_example}. In this table, we use sensor and actuator data from the first stage of the SWaT testbed. In this testbed, sensors include a flow indicator transmitter (FIT101) and level indicator transmitter (LIT101), while actuators consist of pumps (P101, P102) and a moving valve (MV101). Sensor values are continuous, and actuator values are discrete (0 for opening; 1 for opened; 2 for closed). Formally, we use $u$ to denote an observation at a time point, consisting of sensor and actuator values, as shown in equation \ref{eq:system_state}:
\begin{equation}
\label{eq:system_state}
    u=[u_{s1},u_{s2},u_{s3},...u_{a1},u_{a2},u_{a3}, ...],
\end{equation}
where $u_{sj}$ represents a value for the $j^{th}$ sensor and $u_{ak}$ represents the value for the $k^{th}$ actuator. We define $u^i\in R^n$ to be the system state at the $i^{th}$ time point, and $U^i$ to be the sequence of states before the $i^{th}$ time point.

\begin{table}[htb]
\centering
\small
\setlength\tabcolsep{4pt}
\caption{Running Example}
\label{tab:running_example}
\begin{tabular}{|c|c|c|c|c|c|c|c|c|}
\hline
 \textbf{Timestamp} & \textbf{FIT101} & \textbf{LIT101} & \textbf{MV101} & \textbf{P101} &\textbf{P102} & \textbf{Label} & \textbf{Difficulty} & \textbf{\# of Batch}\\ \hline
10:00:00 &	2.43 &	522.84&	2&	2&	1 &Normal & 0.1& 1\\ \hline
10:00:01 & 2.45&	522.88&	2&	2&	1 & Normal & 0.1& 1\\ \hline
...&...&...&...&...&...&...&...&...\\ \hline 
 10:29:13 &2.44&	816.84 &2 &	1 &	1& Normal & 0.4 & 23\\ \hline
\cellcolor{Gray}10:29:14 & \cellcolor{Gray}2.49 &	\cellcolor{Gray}817.67&	\cellcolor{Gray}23&	\cellcolor{Gray}1&	\cellcolor{Gray}1 & \cellcolor{Gray}Attack &\cellcolor{Gray} 0.5&\cellcolor{Gray} 23\\ \hline
\cellcolor{Gray}10:29:15 & \cellcolor{Gray}2.54	&  \cellcolor{Gray}817.94	& \cellcolor{Gray}23& 	\cellcolor{Gray}1	& \cellcolor{Gray}1 & \cellcolor{Gray}Attack &\cellcolor{Gray} 0.5&\cellcolor{Gray} 23\\ \hline 
\cellcolor{Gray}...&\cellcolor{Gray}...&\cellcolor{Gray}...&\cellcolor{Gray}...&\cellcolor{Gray}...&\cellcolor{Gray}...&\cellcolor{Gray}...&\cellcolor{Gray}...&\cellcolor{Gray}...\\ \hline
\cellcolor{Gray}10:44:53 &	\cellcolor{Gray}6e-4	& \cellcolor{Gray}869.72	& \cellcolor{Gray}1 &	\cellcolor{Gray}2 &	\cellcolor{Gray}1 & \cellcolor{Gray}Attack
&\cellcolor{Gray} 0.4&\cellcolor{Gray} 22\\ \hline
\end{tabular}
\end{table}

An attack against MV101 is also shown in this table. The system starts at 10:00:00, pumping water from outside into the tank. As the water level (LIT101) in the tank grows above the upper limit, MV101 should be closed to prevent a water overflow. However, an attacker forces MV101 to be opened regardless of the readings from LIT101. This attack starts at 10:29:14 and ends at 10:44:53, causing real water overflow damage to the plant. Our objective is to detect such attacks in advance before they cause actual damages, with the help of $U^i$. \textcolor{black}{We aim to make both coarse-grained and fine-grained predictions. For coarse-grain predictions, we calculate the number of attacks detected. For fine-grained predictions, we predict the labels for each data sample. For instance, we want to predict the label column shown in Table \ref{tab:running_example} for each given time step.} 

We also show the assigned difficulty score and batch number for CL. Both of them are calculated with \method, which we will introduce in detail in Section \ref{sec:approach}. Intuitively, we define difficulty score to be a number between 0.0 and 1.0, indicating the level of difficulty this sample presents. To be more specific, the higher the difficulty score is, the more difficult this sample is. \textcolor{black}{We assign a new batch number for each sample and change the order based on these scores. Formally, we use $d_i$ and $b_i$ to denote the difficulty score and batch number for sample $i$. For instance, the first sample in Table~\ref{tab:running_example} has a difficulty score of 0.1 ($d_1=0.1$) and its batch number is 1 ($b_1=1$). This means that the first sample is relatively easy for the model to learn, and it will be included in the first batch of the training dataset. In this work, we aim to predict the label column at each time point.}

\section{Approach} \label{sec:approach}
\begin{figure*}[ht]
\includegraphics[width=0.70\textwidth]{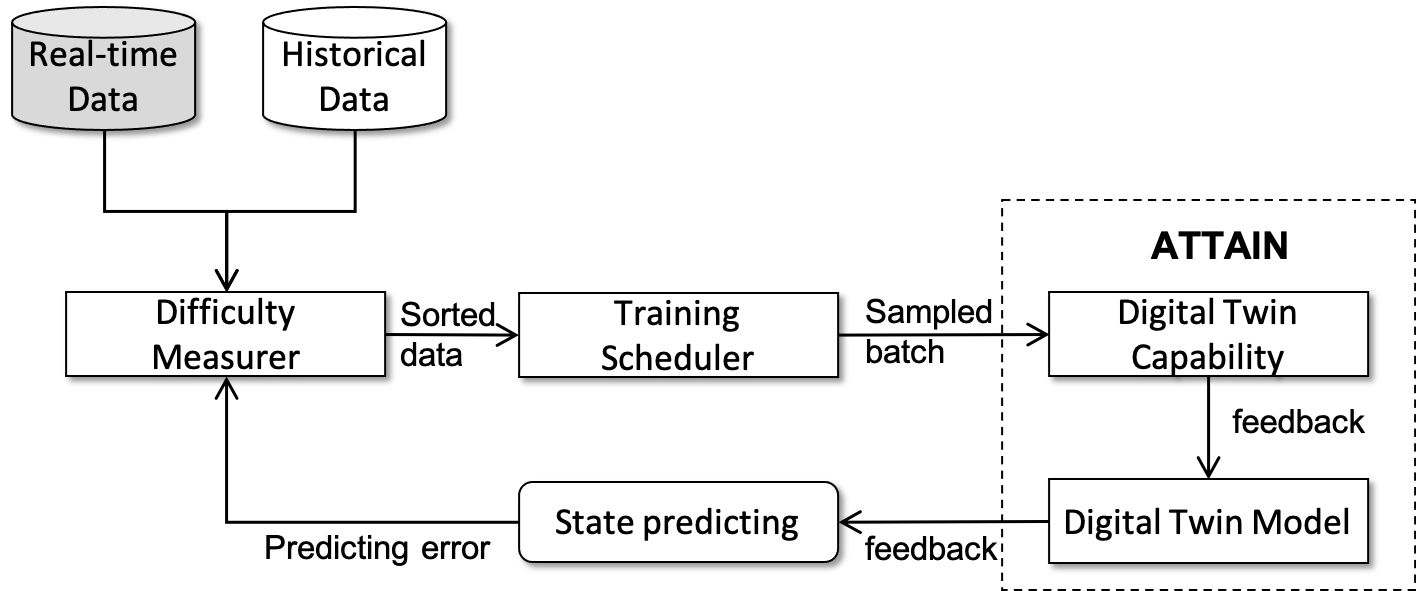}
\centering
\caption{Overview of LATTICE}
\label{fig:overview}
\vspace{-10pt}
\end{figure*}

\method follows the general CL framework proposed in~\cite{Wang}. As we discussed in Section \ref{sec:background}, the main idea of CL is about training models from easier data to harder data. Therefore, a general CL design consists of \textit{difficulty measurer} and \textit{training scheduler}. Difficulty measurer decides the relative "difficulty" of each sample, while \textit{training scheduler} arranges the sequence of data subsets throughout the training process based on the judgment of \textit{difficulty measurer}. As shown in Figure \ref{fig:overview}, all the training examples are sorted by the \textit{difficulty measurer} from the easiest to the hardest and passed to the \textit{Training Scheduler}. Then, at each training epoch $t$, the \textit{training scheduler} samples a batch of training data from the relatively easier examples and sends it to the \textit{ATTAIN} for training. With progressing training epochs, the \textit{Training Scheduler} decides when to sample from harder data. As shown in the running example (Table \ref{tab:running_example}), the Difficulty column presents the difficulty scores given by the \textit{difficulty measurer}, indicating the relative "difficulty" of each sample. For instance, the sample data at 10:00:00 is assigned a difficulty score of 0.9, i.e, $score(u^{10:00:00})=0.9$, while the difficulty score of the sample at 10:29:12 is 0.5, i.e, $score(u^{10:29:12})=0.5$. This tells that sample $u^{10:00:00}$ is relatively harder for the model to learn. With these difficulty scores, the training scheduler decides which samples should be included in each batch. The general principle is that easy samples should be included first. After calculation, the training scheduler assigns new batch numbers for $u^{10:00:00}$ (batch number=23) and $u^{10:19:12}$ (batch number=2). In the following section, we will present more details about the \textit{difficulty measurer}, \textit{training scheduler} and the extension to ATTAIN in Section \ref{subsec:dm}, Section \ref{subsec:ts}, and Section \ref{subsec:ae}, respectively. 
  
\subsection{Difficulty Measurer}\label{subsec:dm}
\textcolor{black}{Wang et al.~\cite{Wang} pointed out that it is difficult to find the best combination of difficulty scorer and training scheduler for a specific task except for performing an exhaustive search, which is often impossible to do for a complex problem. Thus, we did not perform an exhaustive search. However, we investigated various options of difficulty scorers and selected the best-performed ones.} Particularly, we use two different types of difficulty measurers, namely predefined and automatic difficulty measurers. Predefined difficulty measurers are designed based on human prior knowledge with no data-driven models or algorithms involved, while automatic difficulty measures are learned by data-driven models or algorithms. Predefined difficulty measurers have been proven to be effective in various tasks and have been popularly used in multiple domains due to their simplicity~\cite{Wang} (also see Section \ref{subsec:CL}). However, automatic difficulty measurers require less expert knowledge and can interact with current models to adapt to new data.  

\subsubsection{Predefined Difficulty Measurer}\label{subsubsec:pdm}
\textcolor{black}{
Predefined difficulty measurers are usually designed with the help of domain knowledge. Researchers have manually designed various difficulty measurers based on data characteristics of specific tasks~\cite{Penha,ranjan2017curriculum,Eyre2020,Spitkovsky2010,simple}. We adapt these works to CPS data and propose our own predefined difficulty measurers $s_{pdm}$. As shown in Equation \ref{eq:pdm}, \textcolor{black}{$s_{pdm}$ is calculated with four different types of domain knowledge: complexity ($s_{comp}$), diversity ($s_{div}$), noise ($s_{noi}$), and system vulnerability ($s_{vul}$). Each of them is normalized and then all are summed.}
\begin{equation} 
    \label{eq:pdm}
    s_{pdm}=normalize(s_{comp})+normalize(s_{div})+normalize(s_{noi})+normalize(s_{vul})
\end{equation}
\textbf{Complexity} stands for the structural complexity of data samples, such as the dimension and input space of data. Most of the time, the number of sensors and actuators remains unchanged during the operation of CPS, while attackers can potentially change sensor and actuator values into new ones. This increases the input size of data samples, which subsequently increases the complexity. Therefore, we calculate the complexity of each data sample as in Equation \ref{eq:complex}}
\begin{equation}
    \label{eq:comp}
    s_{comp}= size(sensors\; and\; actuators\; input\; space)
\end{equation}

\noindent\textcolor{black}{
\noindent\textbf{Diversity} stands for the distributional diversity of data samples. A data sample is considered to increase the diversity of the dataset if the frequency of this data sample is low. A higher diversity potentially indicates that data samples are distributed in more types. In this paper, the diversity difficulty measurer adopts the naive Bayesian assumption and calculates the probability of this data sample as in Equation \ref{eq:diversity}. Under this assumption, we consider actuators to be independent of each other and use frequency to estimate this probability. The probability of this data sample is factorized as the product of the frequency of each actuator value. 
    \begin{equation}
       \label{eq:diversity}
       s_{div}= \prod_{i=0}^{n}frequency(u_i)
    \end{equation}
}

\noindent\textcolor{black}{
\textbf{Noise} is about the noise level of data samples. A data sample tends to be noisy if its value deviates from the context. The noise difficulty measurer calculates this deviation as the standard score, as shown in Equation \ref{eq:noise}, where $\mu$ and $\sigma$ denote the expected value and standard deviation of data samples from the context.
    \begin{equation}
    \label{eq:noise}
        s_{noi}= \frac{u_i-\mu_i}{\sigma}
    \end{equation}  
}

\noindent\textcolor{black}{
\textbf{Vulnerability} determines how vulnerable the system is to attacks. We define vulnerability as the distance to a known attack since attackers tend to invade the system when it is vulnerable and we hypothesize that samples around time points of attacks are more vulnerable to attacks. Consequently, samples with higher vulnerability should be assigned higher difficulty scores because of the increased complexity, diversity, and noise. We argue that all these three characteristics increase when attacks happen or are about to happen, as explained below: 
\begin{itemize}
    \item \textbf{Complexity}. Typically during attacks, values of sensors and actuators become unstable and often deviate from normal ranges. This increases the input space while the input dimension remains the same (the sum of sensors and actuators). We assume that samples with larger input space are more complicated to be trained on. 
    \item \textbf{Diversity}. This increase comes from new data patterns introduced by attackers, which tend to manipulate the values of actuators and sensors for their intended purposes. Such a manipulation inevitably increases the diversity of data, thereby increasing the difficulty of model training on this data. 
    \item \textbf{Noise}. Given that our approach relies on DTM to give ground truth labels of real-time data, the confidence level of this labeling process is lowered when attacks happen. In other words, labels given by DTM become noisy. Many deep learning models can be induced to fit these noisy label distributions instead of real data distribution. Subsequently, these models tend to make inaccurate predictions around time points of attacks, hence the high difficulty score.
\end{itemize}
In the context of our running example in Section \ref{sec:runningexample}, an attack was introduced at 10:29:14, which implies that the complexity, diversity, and noise of samples around $10:29:14$ tend to increase. As a result, samples around $u^{10:19:14}$ are hard to be trained on and hence obtain high difficulty scores.
}

\textcolor{black}{
We formally define a distance-based vulnerability difficulty measurer $s_{vul}$ based on the above-discussed hypothesis as shown in Equation \ref{eq:vul}: $s_{vul}$ is the time distance $d^i$ between current sample $u^i$ and closest attack sample $u^c$. As in the running example, the closest attack to sample $u^{10:00:00}$ is $u^{10:29:14}$. Therefore, $s_{vul}$ is calculated as $d^i=|timeDistance(u^{10:29:14}-u^{10:00:00})|$. To enable calculations with other difficulty measures in the future, we further scale it to be within $0$ and $1$ with the min-max normalizer as in Equation \ref{eq:pds}, where $min$ and $max$ denote the minimum and maximum values of time distance in the training samples, respectively. Finally, we use this scaled distance as difficulty score $s_{vul}^i$ for sample $u^i$ as in Equation \ref{eq:pds}:
    \begin{equation}
    \label{eq:vul}
        d^i=|timeDistance(u^c-u^i)|
    \end{equation}
    \begin{equation}
        \label{eq:pds}
        s_{vul}=\frac{d^i-min}{max-min}
    \end{equation}
Noticeably, the $timeDistance$ function should be carefully designed to preserve temporal characteristics of anomaly detection data, which is usually in the form of sequences. To that end, we use a sliding window mechanism for this function. In particular, we define hyperparameter $s_{window}$ to denote the size of this sliding window, within which samples share the same distance. Let $d^i_{o}$ and $d^c_o$ to be the original time distances for sample $u^i$ and sample $u_c$. We calculate the window distance for any given sample as in Equation \ref{eq:timeDist}: 
\begin{equation}
    \label{eq:timeDist}
    timeDistance(u^i)=floor(\frac{ d_o^i-d_o^c}{s_{window}})
\end{equation}
}

\subsubsection{Automatic Difficulty Measurer}\label{subsubsec:adm}
Despite the simplicity and effectiveness of predefined difficulty measurers, they have some essential limitations~\cite{Wang}. First, a predefined difficulty measurer remains unchanged during runtime, i.e., being unable to adapt to new data generated from CPS in operation. Second, a predefined difficulty measurer requires a good grasp of domain knowledge, which can be quite expensive and time-consuming in practice. Last but not least, the definition of difficulty for humans and machines can be quite different; what humans assume to be easy can be quite difficult for machines to comprehend. This discrepancy of decision boundaries between humans and machines causes challenges for experts to define difficulty scores manually.

To alleviate these problems, various automatic difficulty measurers have been developed and explored in the literature, including self-paced learning~\cite{Kumar}, transfer teacher~\cite{Carpuat2017}, reinforcement learning teacher~\cite{Graves2016}, and other automatic difficulty measurers~\cite{Tsvetkov,Jiang2018,294207}. Inspired by these methods, we modify difficulty scores automatically with prediction errors, which are critical indicators of CPS uncertainty. Substantial work has been conducted in the literature, demonstrating the importance of handling uncertainties in CPS security and safety~\cite{liping,liping-tosem,ma2019testing,zhang2019uncertainty,qinghua-fse,zhang2019modeling}. In our context, we focus on prediction errors of DTM, which is pretrained on historical data. DTM in ATTAIN simulates corresponding CPS with high realism. Therefore, higher prediction errors of DTM indicate higher noise levels of labels produced by DTM. As mentioned in Section \ref{subsubsec:pdm}, training deep learning models with noisy data is more difficult. Therefore, we assume samples with higher prediction errors should be assigned with higher difficulty scores. Based on this assumption, we define the following two types of automatic difficulty measurers: Hamming Distance-based Measurer (HDM, Definition \ref{def:hdm}) and Cross Entropy-based Measurer (CEM, Definition \ref{def:cem}). Hamming distance is commonly used to calculate the difference of two strings of equal length, while cross-entropy loss estimates uncertainty by comparing real distribution and prediction. We are aware that there are other distance-based and entropy-based metrics, we, however, argue that hamming distance and cross-entropy are commonly used and representative. In the future, we will explore other options.
\begin{definition}\label{def:hdm}
    HDM. 
    
    Let $\hat{u}_{DTM}$ be the predicted state for the current time point. We define HDM score $s_{HDM}$ as in equation \ref{eq:hdm}:
    \begin{equation}
        \label{eq:hdm}
        s^i_{HDM}=\frac{\sum_{k=1}^n{\hat{u}^i_k+u^i_k}}{n}
    \end{equation}
    where $\hat{u}^i_k$ denotes the $k$th element of predicted state vector at time point $i$. As in the running example, $u^{10:00:00}_3$ denotes the third element of sample $u^{10:00:00}$, which is the value of MV101 (2). 
\end{definition}
\begin{definition}\label{def:cem}

CEM.

We define CEM score $s_{CEM}$ as in equation \ref{eq:cem}:
    \begin{equation}
        \label{eq:cem} 
        s^i_{CEM}=-\sum_{k=1}^n {softmax(u)^i_k\times log(softmax(\hat{u})^i_k)}
    \end{equation}
\end{definition}

\subsubsection{Combined Difficulty Measurer}\label{subsubsec:cdm}
As previously mentioned in Section \ref{subsubsec:pdm} and Section \ref{subsubsec:adm}, predefined and automatic difficulty measurers have both advantages and disadvantages. Predefined difficulty measurers can incorporate expert knowledge into learning processes, but it is costly in practice. On the other hand, automatic difficulty measurers can self-adjust during training, which saves considerable time and cost. However, automatic difficulty measurers depend only on training losses and prediction errors acquired from data and discard expert knowledge. Therefore, we propose to combine these two paradigms by introducing prior expert knowledge into the automatic learning process.

We define a hyperparameter $\lambda$ ($0<\lambda<1$) to control the influence level of the predefined difficulty measurer (prior knowledge). The combined measurer is defined as in equation \ref{eq:cb1} and equation \ref{eq:cb2}:
\begin{equation}
    \label{eq:cb1}
    s^i_{cb1}=s^i_{HDM}+\lambda s^i_{PDM}
\end{equation}
\begin{equation}
    \label{eq:cb2}
    s^i_{cb2}=s^i_{CEM}+\lambda s^i_{PDM}
\end{equation}

\subsection{Training Scheduler}\label{subsec:ts}
\textcolor{black}{As for the training scheduler, we also explored several options including baby steps~\cite{Platanios2019}, one pass~\cite{Platanios2019}, and root function continuous scheduler~\cite{choi2019pseudo}. However, we investigated the performance of these training schedulers. Results from this investigation show the superiority of the combination of the proposed difficulty scorer and baby step training scheduler.} Baby step scheduler first distributes the sorted data into buckets from easy to hard and starts training with the easiest bucket. After a fixed number of training epochs or convergence, the next bucket is merged into the training subset. Finally, after all the buckets are merged and used, the whole training process either stops or continues with several extra epochs. Note that at each epoch, the scheduler usually shuffles both the current buckets and the data in each bucket and then samples mini-batches for training (instead of using all data at once).

\begin{algorithm*}
\SetAlgoLined
\KwIn{$D$: training dataset; $C$: the difficulty measurer}
\KwOut{$M^*$: the optimal model.}
$D'=sort(D,C)$\;
${D^1,D^2,...,D^k}=D' where\ c(d_a)<c(d_b),d_a\in D^i,d_b \in D^j, \forall{i}<j$\;
$D^{train}=\emptyset$\;
\For{s=1...k}{
$D^{train}=D^{train}\cup D^s$\;
\While{not converged for p epochs}{
$train(M,D^{train})$
}
}
 \caption{Baby Step Training Scheduler}
\end{algorithm*}

\subsection{ATTAIN Extension}\label{subsec:ae}

\begin{figure*}[ht]
\includegraphics[width=0.70\textwidth]{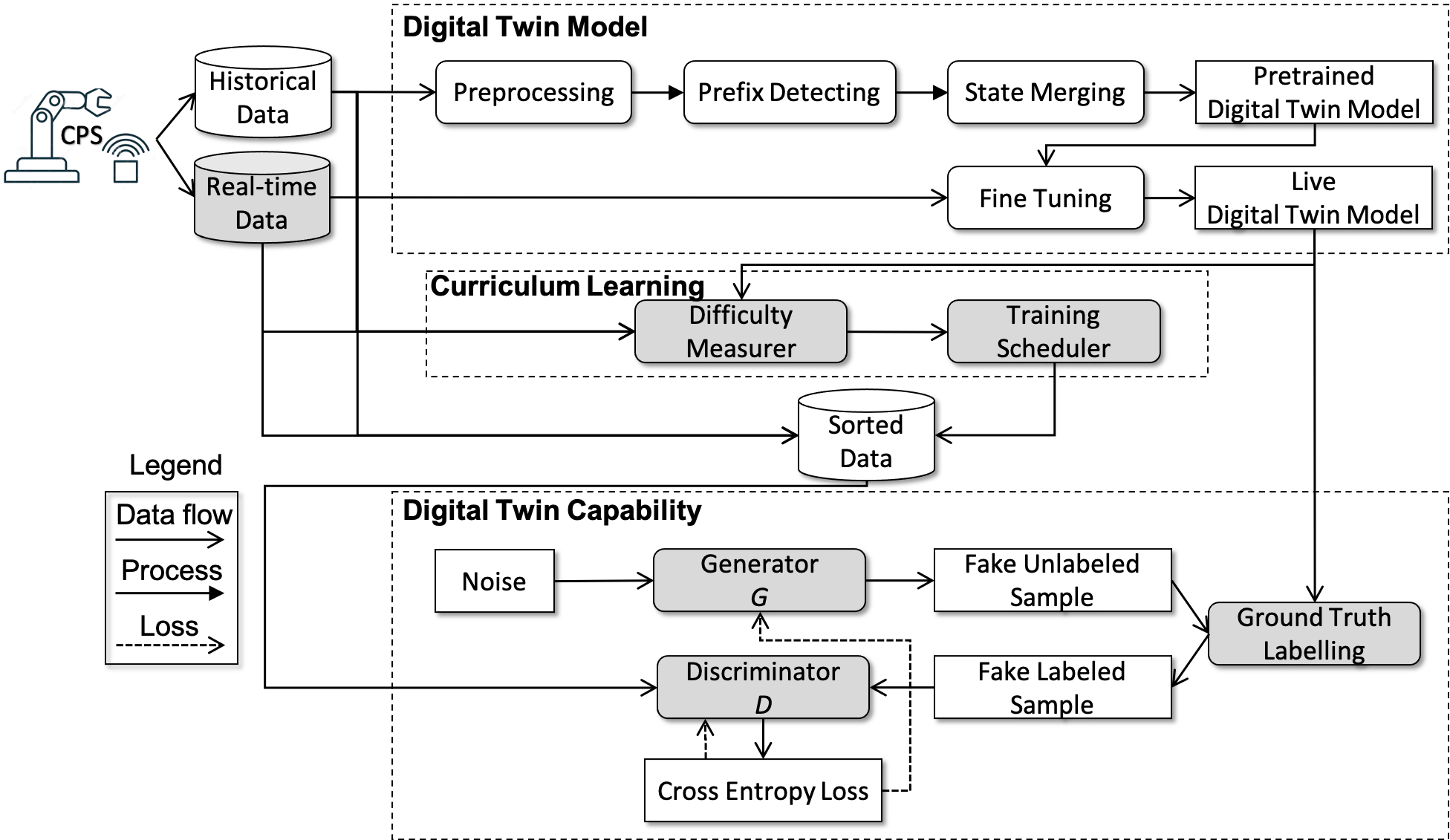}
\centering
\caption{Details of ATTAIN Extension}
\label{fig:attain}
\vspace{-10pt}
\end{figure*}

In addition to implementing CL, we also extended ATTAIN by introducing gated GCN. The top part of Fig.~\ref{fig:attain} shows the overview of ATTAIN. Two types of data need to be collected before training: sensor and actuator values. Data is acquired both from the past, (i.e., \textit{Historical Data}) and in real-time (i.e., \textit{Real-time Data}) from an operational CPS.

Learning digital twin models is a standard state prediction process based on historical and real-time data. It is initially learned as a timed automaton machine from the historical data statically. Though this pre-trained digital twin model tends to simulate the real CPS with high realism, it is limited to capturing only known behaviors. Therefore, this model needs to be further improved with real-time data, allowing it to evolve along with its physical counterpart at runtime. Both the pre-training and online training processes utilize the OTALA~\cite{Maier2014} algorithm, a multi-stage algorithm with the following steps: pre-processing, prefix detection, and state merging. We extend this algorithm to learn probabilistic real-time automaton.  

In this case, the DTC, i.e., the anomaly detector, is trained only on real-time data, where GAN is used as the backbone framework. GAN consists of a generator and a discriminator. The generator learns to generate adversarial samples, mapping from a latent space to input data distribution. The discriminator learns to distinguish among real attack, real normal, adversarial normal, and adversarial attack samples. The anomaly detector is trained in a supervised learning fashion, yet unlabeled adversarial samples are also utilized. To that end, our method calculates a ground truth label as training signals by comparing real sensor and actuator values and predicted values generated by the digital twin model. This ground truth label is then used for the cross entropy loss calculation. Backpropagation is applied in both the generator and discriminator such that the generator produces better adversarial samples while the discriminator becomes more skilled at flagging samples from different categories.

\subsubsection{Digital Twin Model Generation} \label{subsubsec:dtmodel}
A digital twin model is often a behavioral model. Such a model is usually represented as a state chart or a finite automaton and is mostly created manually by domain experts ~\cite{Schiff2011,Ma2011}. However, there exist approaches on the automated construction of Deterministic Finite automaton such as the ones reported in ~\cite{Maier2014,Aichernig2020}. Along with this line, we propose to use Timed Automaton to represent our automatically generated digital twin model. In the automaton theory, a timed automaton is a finite automaton extended with a finite set of real-valued clocks, which can better model real-time systems.

 A Timed Automaton is a tuple $A = (U, T, \delta)$, where,
\begin{itemize} 
    \item $U$ is a finite set of states. In our context, each state $u \subseteq U$ is defined as an observation in a time point, which is a vector consisting of sensor values and actuator values $u=[u_{s1},u_{s2},u_{s3},...u_{a1},u_{a2},u_{a3},...]$;
    \item $T$ is a finite set of transitions, where $T\subseteq U \times U$. For example, for a transition $<u, u'>$, $u, u'\in U$ are the source and destination states; and
    \item $\delta$ is a transition timing constraint $\delta:T\xrightarrow{}I$, where $I$ is a set of probability distribution functions with regard to time, which denotes how much time is needed for a certain transition.
\end{itemize}

We use OTALA as a learning algorithm, which was introduced by Maier et. al.~\cite{Maier2014} as a generic algorithm. We further extend it for our purpose. According to Maier's definition, a transition is triggered by events and finished when the timing constraint clock runs up. However, in the case of CPS, events are not observable and transitions are not deterministic. Therefore, we extend OTALA by dropping the concept of events and introducing probability distribution functions as time constraints. In our example (Table \ref{tab:running_example}), the state vector at 10:00:00 is $u_0=[2.43, 522.84, 2, 2,1 ]$. And at 10:00:01, we observe a new state $u_1=[2.45, 522.88, 2, 2, 1]$ which has not been observed before. Therefore we add $u_1$ into the states set $U$ and add one transition from $u_0$ to $u_1$ into the transition set $T$. Then the algorithm continues to take another input at the next second.

\subsubsection{Digital Twin Capability} \label{subsec:capability}
The core part of Digital Twin Capability is a Generative Adversarial Network (GAN). GAN's capability of producing adversarial samples significantly increases the volume of training data. It consists of two independent models: generator $(G)$ and discriminator $(D)$. $G$ learns to generate more realistic samples while $D$ continuously improves its capability of distinguishing adversarial samples from real samples (as also discussed in Section \ref{subsec:bg_gan}). Here we modify this vanilla GAN for our purpose. We calculate a ground truth label with the help of the digital twin model and denote each sample with more fine-grained labels: real normal, real attack, adversarial normal, and adversarial attack. In this case, the discriminator is a 4-category classifier, which is trained with much more data compared to most existing anomaly detectors.

In this paper, we extend ATTAIN by replacing GCN in the generator with Gated GCN, while the discriminator stays unchanged. Gating mechanisms have been effective in RNNs such as Gated recurrent units and LSTM. Gating mechanisms have been proven to be effective in fitting more complicated data. They control the information flow through their recurrent cells. In the case of GCN, these gated units control the domain information that flows to pooling layers. The model must be robust to change in domain knowledge and should be able to generalize well across different domains. For instance, in our example (Table \ref{tab:running_example}), LIT101, FIT101, and P101 are connected. ATTAIN requires domain knowledge of SWaT to acknowledge this connection, while LATTICE can learn this during training automatically. In the following part of this section, we elaborate on the structure of the generator.


As described in the running example, let $u[i]$ be the $i$th sensor or actuator values of the system. We consider values within a predefined time window as input, e.g., $u[i],u[i-1],...,u[i-window\_size]$. To get a spatial representation $spatial[i]$ for each time step, first, the input is fed into a multi-layer network, consisting of an \textit{Input Layer}, a \textit{GCN Layer}, and a \textit{Pooling Layer}. Second, the spatial representation of all time steps within the window is fed into an \textit{LSTM module} to learn temporal characteristics. The final output of the generator is adversarial samples $fake[i]$ containing both temporal and spatial characteristics.  
\begin{itemize}
    \item \textbf{Input Layer.} Let $u[i]$ be the raw input values in the $i$th time point, consisting of actuator values $a[i]$ and sensor values $s[i]$, which are discrete values and continuous values, respectively. For discrete values, we encode them into one-hot vectors as shown in Equation \ref{eq:discrete}, while for continuous values, we expand them to 3-dimensional vectors, adding their upper limits and lower limits as two additional dimensions as shown in Equation \ref{eq:continuous}. $u[i]$ is made up by concatenating $a[i]$ and $s[i]$ as in equation \ref{eq:io_vec}:  
    \begin{equation}\label{eq:discrete}  
        a'[i]=oneHot(a[i])
    \end{equation}
    \begin{equation}\label{eq:continuous}
        s'[i]=concat(s[i],upper\_limit,lower\_limit)
    \end{equation} 
    \begin{equation}
        u'[i]=concat(a'[i],s'[i]) \label{eq:io_vec}   
    \end{equation} 
    
    \item \textbf{GCN Layer.} GCN Layer captures interdependent relationships among sensors and actuators. In ATTAIN, GCN takes a specification graph as well as sensor and actuator values as the input. In this graph, each actuator and sensor is viewed as a node, while an edge is drawn when there is a connection between these two nodes according to the CPS process. However, specification graphs are not always available, which motivates us to extend GCN with Gated GCN to learn graph edges automatically. As shown in the running example, we find the connection between MV101, P101, LIT101, and FIT101 in the specification graph. Therefore the corresponding edge weight of these four nodes will be set as 1 in ATTAIN, while other edges such as edges between P101 and P102 will be set as 0. However, in LATTICE, the weight matrix will be initialized randomly at the beginning and updated during training. We formally define this gated GCN Layer as follows. 
    
    Let $P[i]$ be the graph edge weight matrix at time point $i$. We first calculate a new candidate matrix $P_c[i]$ as in equation \ref{eq:cand}, where $W_c$ and $b_c$ are weight and bias matrices, respectively.  
    \begin{equation}
        \label{eq:cand}
        P_c[i]=tanh(P[i]W_c+b_c) 
    \end{equation}
    We then calculate the control matrix $P_w$ as in equation \ref{eq:control}:
    \begin{equation}
        \label{eq:control}
        P_w[i]=sigmoid(P[i]W_w+b_w)
    \end{equation}
    Finally, we update edge matrix $E$ for the current time point as in equation: \ref{eq:edge}
    \begin{equation}
        \label{eq:edge}
        E=P_w[i] \times P_c[i]
    \end{equation} 
    Consequently, we have an updated graph $g$ as in equation \ref{eq:graph}. Gated GCN takes graph $g$ and vector $u'[i]$ as input as shown in equation \ref{eq:GCN}:
    \begin{equation}\label{eq:graph}
        g=(E,V)
    \end{equation}
    \begin{equation}
        gcn_{out}[i]=GatedGCN(g,u'[i])\label{eq:GCN}
    \end{equation}
    
    \item \textbf{Pooling Layer.} In the pooling layer, all the GCN outputs collapse into one vector, as shown in equation \ref{eq:pooling} below:
    \begin{equation}
        spatial[i]=maxPooling(gcn_{out})\label{eq:pooling}
    \end{equation}
    where $spatial[i]$ is the spatial representation vector of timestep $i$.
    
    \item \textbf{LSTM Layer.} In the LSTM Layer, spatial representations from different time steps are concatenated together as input, as shown in Equation \ref{eq:inputw}. LSTM learns temporal features by calculating hidden states between each time step as in Equation \ref{eq:lstm}. The last hidden state $h[i]$ is used as our final representation $fake[i]$ for time step $i$ as shown in equation \ref{eq:fake}:
    \begin{equation} 
    \label{eq:inputw}
        input_w=concat(spatial[i:i-window\_size]) 
    \end{equation}
    \begin{equation}
    \label{eq:lstm}
        h[i]=LSTM(input_w)
    \end{equation}
    \begin{equation}
    \label{eq:fake}
        fake[i]=h[i]
    \end{equation}
\end{itemize}



\section{Experiment Design and Execution}\label{sec:experimentDesign}
Section~\ref{sub:researchquestions} presents the research questions (RQs) that we would like to answer. Section~\ref{sub:casestudies} presents the case studies we used for experimentation, followed by the evaluation metrics (Section~\ref{subsec:evaluationmetrics}). Section~\ref{sub:parameter} provides parameter settings of the experiments, and Section~\ref{sub:expexe} details the experiment execution process. 

\subsection{Research Questions} \label{sub:researchquestions}
In our experiment, we are interested in answering the following four RQs: 
\begin{itemize}
\item \textbf{RQ1:} How effective is our anomaly detector as compared to ATTAIN and the other two baselines from the literature?
\item \textbf{RQ2:} How effective is it for introducing CL in DTC?
\textcolor{black}{
\item \textbf{RQ3:} Is \method on par with the baselines in terms of required training time?
\item \textbf{RQ4:} Is \method on par with the baselines in terms of detection delay time?
}

\end{itemize}
\textcolor{black}{
With RQ1, we aim to compare the effectiveness of our approach with existing approaches from the literature (baselines). RQ2 focuses on evaluating the improvement brought by CL when it comes to DTC. In addition, we perform an ablation study to check the effectiveness of each difficulty measurer inside CL. 
RQ3 is designed to demonstrate the efficiency of \method in terms of training time, whereas RQ4 attempts to show how quickly \method can detect an anomaly.
}

\subsection{Characteristics of the Datasets} \label{sub:casestudies}

We evaluate \method with five CPS datasets, namely Secure Water Treatment (SWaT)~\cite{Mathur2016}, Water Distribution (WADI)~\cite{Ahmed2017}, Battle Of The Attack Detection Algorithms (BATADAL)~\cite{taormina18battle}, PHM challenge 2015 dataset~\cite{phm} and Gas Pipeline Dataset~\cite{Morris2015}. Table \ref{tab:cases} provides key characteristics of these datasets.

\textcolor{black}{
\begin{table}[ht]
\caption{
\textcolor{black}{
Characteristics of the Datasets. $N_{sensor}$, $N_{actuators}$, $N_{train\_attack}$, $N_{test\_attack}$ and $\Bar{N}_{attack\_len}$ denote the number of sensors, number of actuators, number of attacks in the training dataset, number of attacks in the test dataset, \textcolor{black}{and the average length of attacks}, respectively. The total number of samples is $N$.
}
}
\label{tab:cases}
\resizebox{\textwidth}{!}{
\begin{tabular}{|c|c|c|c|c|c|c|}
\hline
Dataset      & $N_{sensor}$ &$N_{actuators}$ & $N_{train\_attack}$ & $N_{test\_attack}$ & $\Bar{N}_{attack\_len}$ & $N$ \\ \hline
SWaT         &25 &26 & 33                     & 8                     & 109                       & 946722           \\ \hline
WADI         &42 &61 & 12                     & 3                     & 665                       & 1221372          \\ \hline
BATADAL      &26 &17 & 5                      & 2                     & 70                        & 12936            \\ \hline
PHM 2015     &4 - 16 & 4& 9284-292064            & 2322-73016            & 2                         & 79916-1152651    \\ \hline
Gas Pipeline &16 &7 & 51011                  & 12753                 & 4                     & 236179           \\ \hline
\end{tabular}
}
\end{table}
}

SWaT is a CPS testbed for water treatment. Its main functionality is about producing filtered water through a series of stages. The testbed consists of 25 sensors and 26 actuators. The SWaT dataset was produced from the testbed during its operation. \textcolor{black}{There are 41 attacks in total. We split these 41 attacks into training (33) and testing (8) datasets. The average length of an attack, i.e., the number of samples in the attack, is 109 in this dataset.} 

The second case study is the WADI data produced from the WADI testbed -- an extension to the SWaT testbed. The main functionality of the WADI testbed is the secure distribution of water. The dataset generated from the WADI testbed has two weeks' normal operation data, whereas it has two days' attack data. \textcolor{black}{
This testbed consists of 42 sensors and 61 actuators. There are 15 attacks in total. We split these 15 attacks into two sets: 12 for training and 3 for testing. The average length of an attack is 665 in this dataset.
} 

The BATADAL dataset is an extension of the WADI dataset. The attacks were designed for an attack detection competition. \textcolor{black}{
This testbed consists of 26 sensors and 17 actuators. There are 7 attacks in total, which are divided into 5 for training and 2 for testing. The average length of an attack is 70.
} 

The PHM challenge 2015 dataset focuses on the operation of plants and the capability to detect failure events of the plants in advance. This dataset is a collection of data from 70 plants and the number of sensors and actuators varies from plant to plant. Therefore, we show the ranges (4-16 sensors and 4 actuators) of this dataset in Table \ref{tab:cases}. \textcolor{black}{
The number of attacks also varies from plant to plant. We split these attacks into two groups: 9284-292064 for training and 2322-72016 for testing. The average length of an attack is around 2, much smaller as compared to SWaT, WADI, and BATADAL.
} 

The gas pipeline dataset was collected from a gas pipeline testbed, which transports gas from one place to another. \textcolor{black}{
There are 63764 attacks in total, which are split into two groups: 51011 for training and 12753 for testing. The average length of an attack is 4.
} 

\subsection{Evaluation Metrics and Statistical Tests} \label{subsec:evaluationmetrics}
\label{subsubsec:metric.eff}

This section presents the evaluation metrics used herein. For RQ1-RQ3, we introduce three metrics (precision, recall, and F1 score) for effectiveness evaluation in Section \ref{subsubsec:metric.eff}. \textcolor{black}{In Section \ref{subsubsec:metric.utt}, we define a novel metric called Unit Training Time (UTT) for answering RQ4. In Section \ref{subsubsec:metric.ddt}, we propose Detection Delay Time (DDT) for answering RQ4. In Section \ref{subsubsec:testing}, we briefly introduce the employed statistical tests.} 

\subsubsection{Metrics for effectiveness}
\label{subsubsec:metric_eff}
\method aims to provide practical information about anomalies to CPS. Such information includes a general indication of whether the target CPS is under attack and detailed information about the attack, e.g., starting time and duration. Correspondingly, we propose two types of metrics for RQ1-RQ3, namely coarse-grained and fine-grained effectiveness metrics.

~\\
\textcolor{black}{
\noindent\textbf{Coarse-grained effectiveness metric.} This metric provides general information about an attack, i.e., the existence of the attack. One practical usage scenario of \method is that it can detect the existence of an attack without much precise information such as the exact starting and ending time for this attack. Inspired by this scenario, we propose \textit{Anomaly Coverage Rate (ACR)} as in Equation \ref{eq:acr}
}

\textcolor{black}{
\begin{equation}
    \label{eq:acr}
    ACR=\frac{N_{detected}}{N_{total}}
\end{equation}
\textcolor{black}{where $N_{detected}$ denotes the number of attacks detected and $N_{total}$ denotes the total number of attacks. We consider an attack detected if half of the attack instances are correctly classified. ACR takes a value between 0 and 1, where a higher value indicates a higher number of attacks are successfully detected, and vice versa.}
}

~\\
\textcolor{black}{
\noindent\textbf{Fine-grained effectiveness metric.} Although the coarse-grained effectiveness metric can assess \method's ability to detect anomalies in general, details of attacks are neglected by ACR. This motivates us to evaluate \method with instance-wise metrics. Given the binary classification nature of anomaly detection, we use three standard classification metrics~\cite{Canizo2019,Kravchik2018,Yao2017}: precision, recall, and F1 score. In our context, precision is the percentage of correctly detected anomaly instances among all the instances that are predicted as anomalies, while recall is the percentage of correctly detected anomaly instances among all the anomaly instances. The F1 score is the harmonic mean of precision and recall.
}

Formally, precision is defined in Equation \ref{eq:precision}:
\begin{equation}
    \label{eq:precision}
    precision=\frac{TP}{TP+FP}
\end{equation}
where TP and FP stand for True Positive and False Positive, respectively. The recall is defined in Equation \ref{eq:recall}:
\begin{equation}
    \label{eq:recall}
    recall=\frac{TP}{TP+FN}
\end{equation}
where FN denotes False Negative. $F_1$ is defined in Equation \ref{eq:F1}:
\begin{equation}
\label{eq:F1}
    F_1= 2\cdot \frac{precision\cdot recall}{precision + recall}
\end{equation}

\subsubsection{Unit Training Time (UTT).}
\label{subsubsec:metric.utt}

\textcolor{black}{
We observe that the training time of \method depends not only on the model itself, but also on the complexity of the CPS. Therefore, to evaluate the model's efficiency, we propose UTT, a CPS complexity-agnostic metric. Specifically, we first identify the complexity-sensitive execution time $tt$ and divide it by the CPS complexity as in Equation \ref{eq:utt}.
\begin{equation}
    \label{eq:utt}
    utt=\frac{tt}{S}
\end{equation}
\textit{Training time $tt$} denotes the required convergence time of the model (Equation~\ref{eq:et}). We assume that the model converges when losses reach the minimum. In other words, losses from two neighboring batches should have minimal discrepancy (smaller than a threshold value $\delta$, i.e. $\delta=1e-4$).
\begin{equation}\label{eq:et}
    tt=time(convergence) - time(start)
\end{equation}
$S$ stands for the complexity of the CPS. Since there is no out-of-box complexity metric for our purpose, we propose the Metric of Complexity (MoC), which estimates the complexity of CPS from three aspects: CPS itself, attack, and dataset, as formalized in Equation \ref{eq:complex}. }

\begin{equation}\label{eq:complex}
    S=S_{CPS}+S_{ATT}+S_{DAT}
\end{equation}

\noindent\textbf{CPS complexity.} CPS complexity consists of static and automatic complexity (Equation \ref{eq:complex_cps}). Static complexity entails a CPS's internal structure, e.g., the number of sensors and actuators. Such structure decides the dimension of \method's input data in a fixed manner. However, not all the components in this structure are active during training. Hence, we propose automatic complexity, which identifies only active components during training.

\begin{equation}\label{eq:complex_cps}
    S_{CPS}=S_{CPS.static}+S_{CPS.automatic}
\end{equation}
\textcolor{black}{
For \textit{static complexity}, we calculate the number of sensors and actuators in the CPS and normalize it as $z_{n\_s/a}$, allowing this metric to be calculated along with other metrics regardless of the scale of CPS. Moreover, we make a crucial observation that value changes in actuators usually induce changes in sensor values. Such correlation indicates dependencies between sensors and actuators. Therefore, we calculate sensors and actuators' ratio  $z_{r\_s/a}$ in the CPS in addition to the absolute number $z_{n\_s/a}$. Finally, we add $z_{r\_s/a}$ and $z_{n\_s/a}$ together as the static complexity of CPS as shown in Equation \ref{eq:cps.static}. 
}
\begin{equation}\label{eq:cps.static}
    S_{CPS.static}=z_{n\_s/a}+z_{r\_s/a}
\end{equation}
However, not all the sensors and actuators are active during the process. As in the running example (Table \ref{tab:running_example}), FIT101, LIT101, MV101, and P101 are only active in Phase 1 of the water treatment process and become inactive in other phases. Therefore, we propose \textit{automatic complexity} for calculating only active sensors and actuators at runtime as in Equation \ref{eq:cps.automatic}. Similar to static complexity, we calculate both the number and ratio of active sensors and actuators. We then add the normalized number $z_{n\_s/a}$ and ratio of  $z_{r\_s/a}$ together as in Equation \ref{eq:cps.automatic}.

\begin{equation}\label{eq:cps.automatic}
    S_{CPS.automatic}=d_{n\_s/a}+d_{r\_s/a}
\end{equation}

\noindent\textbf{Attack complexity.} Attack complexity is computed based on the attack's information, e.g., the number of attacks in total. Formally, we calculate the attack complexity as in Equation \ref{eq:complex_att}, where $z_{n\_n/a}$ denotes the normalized number of attacks; $z_{r\_n/a}$ denotes the ratio of attack and normal samples; and $z_{a\_type}$ denotes the normalized number of different types of attacks. 
\begin{equation}\label{eq:complex_att}
    S_{ATT}=z_{n\_n/a}+z_{r\_n/a}+z_{a\_type}
\end{equation}

\textcolor{black}{
\noindent\textbf{Dataset complexity.} Dataset complexity reflects the scale of a given CPS dataset. Size is a good indicator of the scale. Therefore, we normalize the dataset size as $z_{dat\_size}$ and include it in the computation of dataset complexity. Another scale's indicator that we found is concept drift. Concept drift is a phenomenon where the target variable's statistical properties (e.g., mean/variance values of sensors) change over time~\cite{Wang2015}. As a result, dataset complexity increases in the presence of concept drift. According to Scheffel et al~\cite{Scheffel2019}, this phenomenon is prevalent in the CPS domain, which motivates us to include concept drift in dataset complexity computation. To calculate concept drift, we take inspiration from Adaptive Sliding Window (ADWIN)~\cite{Du2014} and propose a novel concept drift complexity metric  $S_{dat\_drift}$. In conclusion, dataset complexity consists of information about dataset size and concept drift as in Equation \ref{eq:complex_dat}.
}

\begin{equation}\label{eq:complex_dat}
    S_{DAT}=z_{dat\_size}+S_{dat\_drift}
\end{equation}  
In detail, concept drift complexity is calculated as in Equation \ref{eq:dat_drift}. $s_p$ and $s_q$ are two random samples selected, respectively from two separate windows $W_i$ and $W_j$ (Equation \ref{eq:sample}). Let sample size be $N$, we calculate the KL divergence of each pair of samples and calculate the average value for all the $N$ pairs.
\begin{equation}\label{eq:dat_drift}
    S_{dat\_drift}= \sum_{p=1,q=1}^{N,N} KL(s_p,s_q)/N 
\end{equation}  
\begin{equation}\label{eq:sample}
     s_p,s_q=sample(W_i), sample(W_j)
\end{equation}

\textcolor{black}{
\subsubsection{Detection delay time (DDT)}
\label{subsubsec:metric.ddt}
As for DDT, we aim to assess how good \method is at detecting anomalies at early stages to prevent further damages. To this end, we count false negative samples $N_{FN}$ at the beginning of an attack which consists of $N$ samples. We then calculate DDT by dividing $N_{FN}$ by $N$ as in
}
Equation \ref{eq:ddt}.
\begin{equation}
    \label{eq:ddt} 
    s_{ddt}=\frac{N_{FN}}{N} 
\end{equation}  
 \textcolor{black}{
DDT indicates the delay of the assessed method when trying to detect an anomaly. It takes a value between 0 and 1. A lower DDT value means that the assessed method can detect attacks at an earlier stage. }

\subsubsection{Statistical Testing} 
\label{subsubsec:testing}

\textcolor{black}{
Due to the inherent randomness of our approach, we employ statistical testing to answer RQ1-RQ4 to determine whether the improvements are statistically significant.
}

\textcolor{black}{
In this paper, we use \textbf{Mann-Whitney U test}, which, unlike $t-$test, makes no assumption on the underlying data distribution.
We test all the pair-wise comparisons in each RQ. In general, to compare \textit{Method A} and \textit{Method B}, we run each method 30 times, as suggested in ~\cite{Arcuri2011}. The null hypothesis is that there is no statistical difference between the two methods. If the null hypothesis is rejected, we conclude that \textit{Method A} and \textit{Method B} are not equivalent.
}

Mann-Whitney U test results reveal the significant difference between \textit{Method A} and \textit{Method B}, whereas the magnitude of this difference is unknown. Such magnitude can be assessed with an effect size. In this paper, we use \textbf{Vargha and Delaney's A12 }, which also requires no knowledge of the underline data distribution. An A12 value ranges from 0 to 1. If $A12=0.5$, it means that the results are obtained by chance. If $A12>0.5$, it means \textit{Method A} has a higher chance of getting better results than \textit{Method B}, and vice versa.

Concretely in this paper, RQ1 and RQ4  involve comparisons between \method (\textit{Method A}) with the baselines (\textit{Method B}). RQ2 involves comparisons between \method (\textit{Method A}) and \latticedtm (\textit{Method B}). RQ3 performs the ablation study by comparing \method (\textit{Method A}) with \pdm, \cem, \hdm,\cemhdm, and \pdmcemhdm (\textit{Method B}).

\subsection{Parameter Settings} \label{sub:parameter}
\textcolor{black}{We use cross-validation to automatically select the hyperparameters of \method. Cross-validation is a commonly used hyperparameter tuning technique in machine learning~\cite{kohavi1995study}. The general idea is to train machine learning models with different parameter settings and find out optimal ones that yield the best performance. Specifically for each hyperparameter set, we split each dataset into training and testing datasets. The training dataset is then further split into ten chunks. We select one chunk as a validation dataset each time, while the remaining nine chunks are used as training datasets. We perform ten such validation processes and calculate an average F1 score as a conclusion. We compare each hyperparameter set's F1 score and find the optimal set based on the comparison result.}

\textcolor{black}{
We divide \method's hyperparameters into two groups: respectively for ATTAIN and CL. We show our optimal hyperparameters found by cross-validation as follows:
\begin{itemize}
    \item \textbf{ATTAIN hyperparameters.} We set the batch size of input data as 64. The hidden dimension of the neural network was set as 100, and Rectified Linear Unit (ReLU) was used as the activation function. As for the GCN layer, we used a gated GCN module and set the number of layers to 2. 
    \item \textbf{CL hyperparameters.} We set the threshold value of the baby step algorithm to 0.8. We use the min-max scaler to calculate the difficulty measurer. Min and max values are calculated with the historical data we have for now. We plan to substitute this with more generic scalers in the future when needed. 
\end{itemize}
}

\subsection{Experiment Execution}\label{sub:expexe}
In this paper, neural network layers were built with the PyTorch framework~\cite{NEURIPS2019-9015}, and the GCN layer was constructed with the PyTorch Geometric (PyG) framework~\cite{Fey/Lenssen/2019}. All the experiments were carried out on Google collaboratory notebooks, with Intel(R) Xeon(R) CPU at 2.00GHz, 12 GB system memory and for GPU, Tesla V100-SXM2 with 16GB memory. To eliminate the effect of randomness, we repeated all the experiments 30 times, and the average results were reported in this paper. For the SWaT dataset, we followed previous works ~\cite{Kravchik2018} by ignoring the first 16000 records because the state of the CPS tends to be highly unstable during that period of time after it is started.      
\section{Results and Analysis} \label{sec:Results}

Section~\ref{subsec:resultsRQ1}-Section ~\ref{subsec:rq4} present the results for RQ1-RQ4, respectively.

\subsection{Results and Analysis for RQ1} \label{subsec:resultsRQ1}

\textcolor{black}{
Table \ref{tab:acr} shows the \textbf{ACR} results of \method and the baselines (i.e., LSTM-CUSUM~\cite{Goh2017}, MAD-GAN~\cite{Li2019}, and ATTAIN~\cite{xudigital}). Both LSTM-CUSUM and MAD-GAN were designed for CPS anomaly detection (see Section~\ref{sec:relatedwork} for more details). ATTAIN is our previous work, which is extended in this paper by introducing CL. We evaluate these methods with ACR on the five public datasets (Section \ref{sub:casestudies}). Anomalies in SWaT, WADI, and BATADAL datasets have a longer duration and fewer occurrences than the PHM 2015 and Gas Pipeline datasets. We can observe that all the methods achieve 100\% on SWaT, WADI, and BATADAL, which means that they never miss any anomaly. Contrarily, ACR values on the PHM 2015 and Gas Pipeline dataset are lower due to the anomalies' shorter duration and higher occurrences. \method achieves the highest ACR values on these two datasets, demonstrating its superior capability at detecting anomalies in a coarse-grained manner.
}

\textcolor{black}{
\begin{table}[ht]
    \centering
    \begin{tabular}{|c|c|c|c|c|}
    \hline
         Dataset        & LSTM-CUSUM  & MAD-GAN & ATTAIN & \method  \\ \hline
         SWaT           & 1           & 1        & 1      & 1       \\ \hline
         WADI           & 1           & 1        & 1      & 1       \\ \hline
         BATADAL        & 1           & 1        & 1      & 1       \\ \hline
         PHM2015        & 0.79        & 0.82     & 0.87   & 0.87     \\ \hline
         Gas Pipeline   & 0.77        & 0.81     & 0.85   & 0.91    \\ \hline
    \end{tabular}
    \caption{\textcolor{black}{ACR results of \method and the baselines} }
    \label{tab:acr}
\end{table}
}

\begin{figure*}[ht]
\includegraphics[width=1\textwidth]{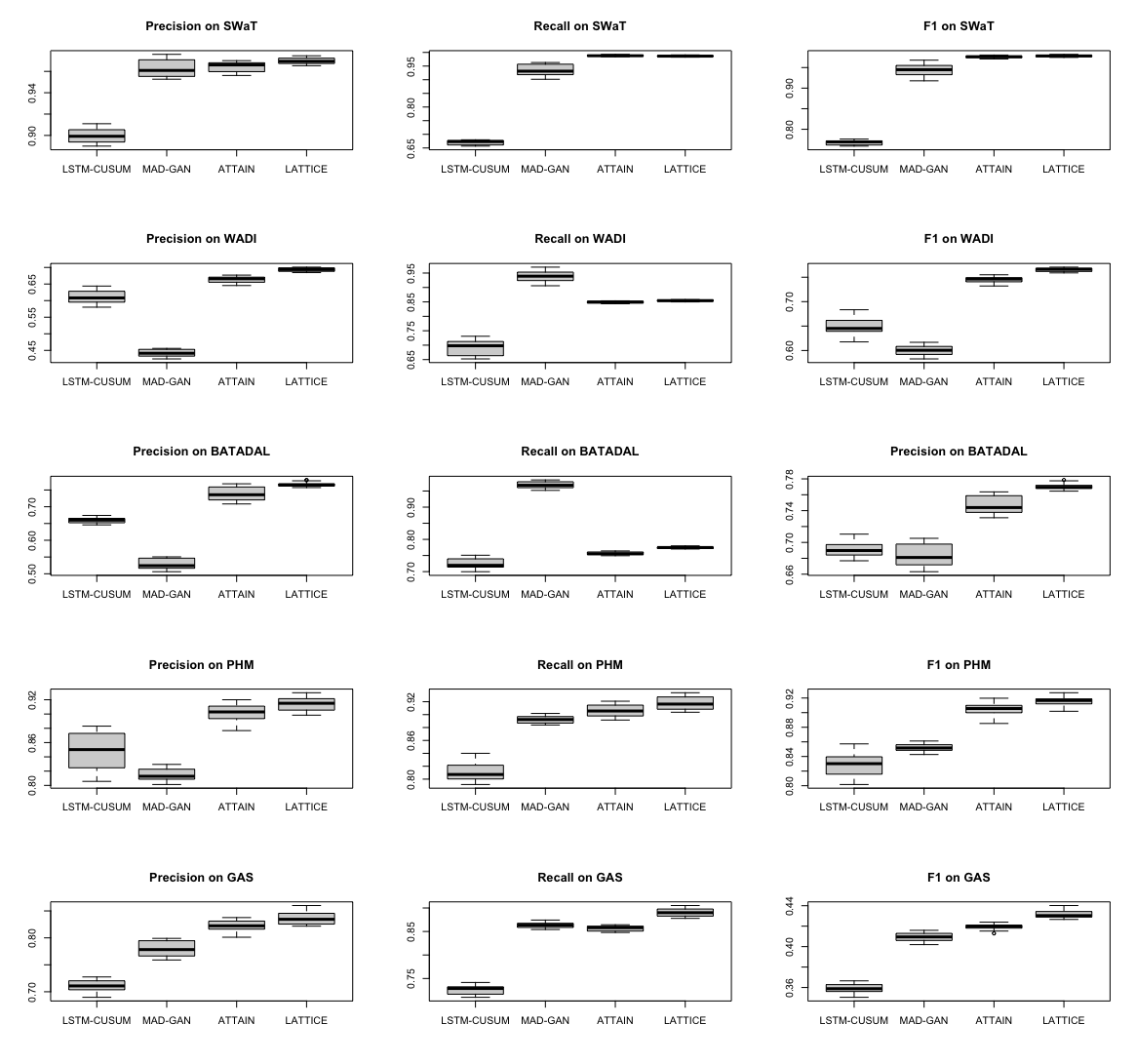}
\centering
\caption{Effectiveness boxplots of \method and the baselines}
\label{fig:boxplot}
\vspace{-10pt}
\end{figure*}

\textcolor{black}{
Figure \ref{fig:boxplot} shows the fine-grained effectiveness metric results of \method. Three boxplots in each row demonstrate the precision, recall, and F1 score on a specific dataset. We can observe that \method outperforms LSTM-CUSUM and MAD-GAN by a large margin except for recall on WADI and BATADAL. \method manages to further improve ATTAIN in all five datasets. For the SWaT dataset, the improvement is slight, but we argue that this is because the precision, recall, and F1 scores are already high, and there is not much room for further improvement at the first place. Also, we can observe that, as compared to the baselines, the results of \method have smaller variances, indicating that \method is more certain about its prediction.
}

\begin{table*}[h]

\centering
\caption{Statistical test results of comparing \method and the baselines}\label{tab:testing_rq1}

\begin{tabular}{|c|c|c|c|c|c|c|}

\hline
Datasets & Metrics& Testing& LSTM-CUSUM & MAD-GAN & ATTAIN \\ 
\hline
\multirow{6}{*}{SWaT}
    & \multirow{2}{*}{Precision}
            & \cellcolor{Gray}p-value & \cellcolor{Gray}<1e-4 & \cellcolor{Gray}9.518e-4 &  \cellcolor{Gray}\textbf{0.1294}\\
        &   &  A12 & 1.0 & 0.701 & 0.846\\
    \cline{2-2}
   
    & \multirow{2}{*}{Recall}
            & \cellcolor{Gray}p-value & \cellcolor{Gray}<1e-4 & \cellcolor{Gray}<1e-4 &  \cellcolor{Gray}\textbf{0.0523}  \\
        &   &  A12 & 1.0 & 1.0 & 0.336 \\
    \cline{2-2}
   
    & \multirow{2}{*}{F1}
            & \cellcolor{Gray}p-value & \cellcolor{Gray}<1e-4 & \cellcolor{Gray}<1e-4 & \cellcolor{Gray}1.13e-2 \\
        &   &  A12 & 1.0 & 1.0 & 0.713 \\
\hline
\multirow{6}{*}{WADI}
    & \multirow{2}{*}{Precision}
            & \cellcolor{Gray}p-value & \cellcolor{Gray}<1e-4 & \cellcolor{Gray}<1e-4 & \cellcolor{Gray}<1e-4  \\
        &   &  A12 & 1.0 & 1.0 & 1.0  \\
    \cline{2-2}
   
    & \multirow{2}{*}{Recall}
            & \cellcolor{Gray}p-value & \cellcolor{Gray}<1e-4 & \cellcolor{Gray}<1e-4 & \cellcolor{Gray}<1e-4 \\  
        &   &  A12 & 1.0 & 0 & 0.875 \\
    \cline{2-2}
   
    & \multirow{2}{*}{F1}
            & \cellcolor{Gray}p-value & \cellcolor{Gray}<1e-4 & \cellcolor{Gray}<1e-4 & \cellcolor{Gray}<1e-4  \\
        &   &  A12 & 1.0 & 1.0 & 1.0  \\
\hline
\multirow{6}{*}{BATADAL}
    & \multirow{2}{*}{Precision}
            & \cellcolor{Gray}p-value & \cellcolor{Gray}<1e-4 & \cellcolor{Gray}<1e-4 & \cellcolor{Gray}1.684e-4 \\
        &   &  A12 & 1.0 & 1.0 & 0.886  \\
    \cline{2-2}
   
    & \multirow{2}{*}{Recall}
            & \cellcolor{Gray}p-value & \cellcolor{Gray}<1e-4 & \cellcolor{Gray}<1e-4 & \cellcolor{Gray}<1e-4  \\
        &   &  A12 & 1.0 & 0 & 1.0 \\
    \cline{2-2}
   
    & \multirow{2}{*}{F1}
            & \cellcolor{Gray}p-value & \cellcolor{Gray}<1e-4 & \cellcolor{Gray}<1e-4 & \cellcolor{Gray}<1e-4  \\
        &   &  A12 & 1.0 & 1.0 & 1.0 \\
\hline
\multirow{6}{*}{PHM2015}
    & \multirow{2}{*}{Precision}
            & \cellcolor{Gray}p-value & \cellcolor{Gray}<1e-4 & \cellcolor{Gray}<1e-4 & \cellcolor{Gray}2.833e-4  \\
        &   &  A12 & 1.0 & 1.0 & 0.791  \\
    \cline{2-2}
   
    & \multirow{2}{*}{Recall}
            & \cellcolor{Gray}p-value & \cellcolor{Gray}<1e-6 & \cellcolor{Gray}<1e-4 & \cellcolor{Gray}1.886e-4  \\
        &   &  A12 & 1.0 & 1.0 & 0.771  \\
    \cline{2-2}
   
    & \multirow{2}{*}{F1}
            & \cellcolor{Gray}p-value & \cellcolor{Gray}<1e-6 & \cellcolor{Gray}<1e-4 & \cellcolor{Gray}3.79e-6  \\
        &   &  A12 & 1.0 & 1.0 & 0.872  \\
\hline
\multirow{6}{*}{Gas Pipeline}
    & \multirow{2}{*}{Precision}
            & \cellcolor{Gray}p-value & \cellcolor{Gray}<1e-4 & \cellcolor{Gray}<1e-4 & \cellcolor{Gray}1.886e-4 \\
        &   &  A12 & 1.0 & 1.0 & 0.811 \\
    \cline{2-2}
   
    & \multirow{2}{*}{Recall}
            & \cellcolor{Gray}p-value & \cellcolor{Gray}<1e-4 & \cellcolor{Gray}<1e-4 & \cellcolor{Gray}<1e-4 \\
        &   &  A12 & 1.0 & 1.0 & 1.0  \\
    \cline{2-2} 
   
    & \multirow{2}{*}{F1}
            & \cellcolor{Gray}p-value & \cellcolor{Gray}<1e-4 & \cellcolor{Gray}<1e-4 & \cellcolor{Gray}<1e-4  \\
        &   &  A12 & 1.0 & 1.0 & 1.0  \\
  
\hline
\end{tabular}
\end{table*} 
 
As mentioned in Section \ref{subsec:evaluationmetrics}, we repeated all the experiments 30 times and performed statistical tests. Table ~\ref{tab:testing_rq1} shows the results of comparing \method with each baseline. 
\textcolor{black}{
We consider $p-value<0.01$ as a significant difference. In terms of precision, \method is significantly better than LSTM-CUSUM (5/5), MAD-GAN (5/5), and ATTAIN ( 4/5 ). The minimum effect size for all the comparisons is 0.701. In terms of recall, \method is significantly better than LSTM-CUSUM (5/5), MAD-GAN  (3/5), and ATTAIN (4/5), while the minimum effect size is 0.771. Regarding the F1 score, \method is significantly better than all the baselines for all the datasets with the minimum effect size of 0.713. 
}

\begin{Summary}{RQ1}{}
\textcolor{black}{
We conclude that \method outperforms the baselines in terms of both the coarse-grained (i.e., ACR) and fine-grained effectiveness metrics. Statistical test results show that this improvement is significant, and \method is more likely to yield better results. In short, our anomaly detector is more effective than the baseline methods in the literature. 
}

\end{Summary}

\subsection{Results and Analysis for RQ2} \label{subsec:resultsRQ2}

\begin{figure*}[ht]
\includegraphics[width=1\textwidth]{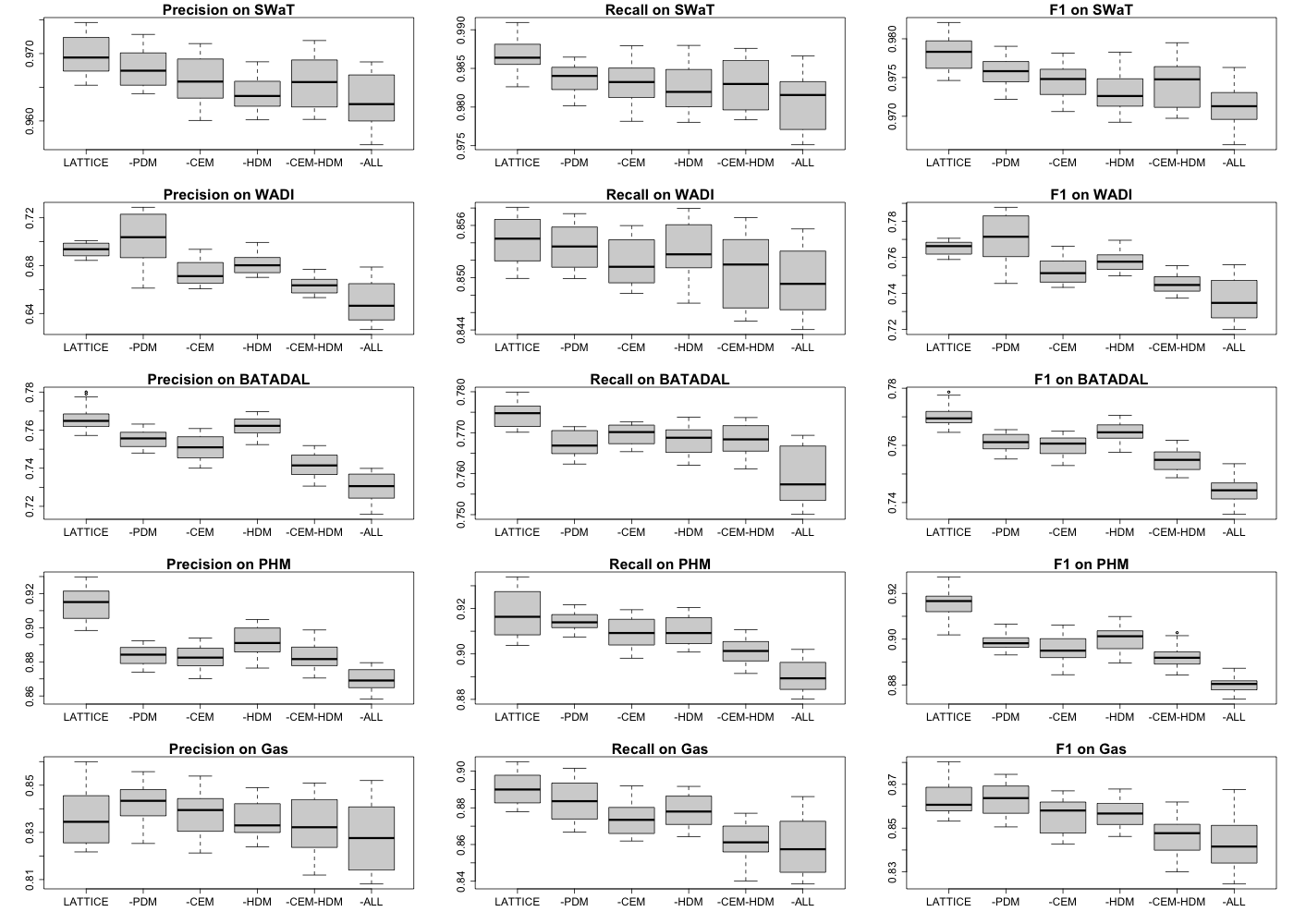}
\centering
\caption{Boxplots of the ablation study experiment results}
\label{fig:ablation}
\vspace{-10pt}
\end{figure*}

\textcolor{black}{
The introduction of CL is the major contribution of \method. To better understand CL's influence, we evaluate the effectiveness of CL and perform an ablation study in this section.
\noindent\textbf{CL effectiveness. }To evaluate CL's effectiveness, we compare \method and \pdmcemhdm, which removes CL from \method. Figure \ref{fig:ablation} presents the box plots of \method and \pdmcemhdm. We see a great performance decline when removing CL from \method. \pdmcemhdm's variance is also larger than \method, indicating higher prediction uncertainty.
}

\textcolor{black}{
We also report the statistical test results in Table \ref{tab:testing_rq2}. The last column compares \method with \pdmcemhdm. CL's improvement on all  five datasets in precision, recall, and F1 score is significant ($p-value<0.05$), except for recall on WADI. The corresponding effect sizes are also strong ($minimum=0.6944$), indicating that \method has a higher probability of yielding better results than \pdmcemhdm.
}

\textcolor{black}{
\noindent\textbf{Ablation study}
In \method, we explore three types of difficulty measurers: PDM, CEM, and HDM. PDM is a predefined difficulty measurer, while CEM and HDM are automatic difficulty measurers focusing on entropy and vector distance, respectively. We argue that each difficulty measurer contributes to the performance improvement of \method. Therefore it is valuable to assess the effectiveness of each difficulty measurer. To that end, we perform an ablation study of each difficulty measurer. 
\begin{itemize}
    \item \textbf{PDM effectiveness.} \pdm removes the PDM difficulty measurer from \method. Figure \ref{fig:ablation} shows that removing PDM decreases recall on all five datasets. The precision on SWaT, BATADAL, and PHM 2015 also decreases while the precision on WADI and Gas Pipeline increases. Mann-whitney U-test results (Table \ref{tab:testing_rq2}) show that the decreases are significant ($p-value<0.05$) except for recall on WADI. The effect size of the F1 score is very strong on SWaT, BATADAL, and PHM 2015 datasets.
    \item \textbf{CEM effectiveness.} \cem removes the CEM difficulty measurer from \method. We can observe from Figure \ref{fig:ablation} that removing CEM decreases precision, recall, and F1 score in all cases. Mann-whitney U-test results demonstrate these decreases' significance ($p-value<0.05$) except for precision on the gas pipeline dataset. The effect sizes are strong with a minimum of 0.7189 (F1 score on gas pipeline dataset).
    \item \textbf{HDM effectiveness.} \hdm removes the HDM difficulty measurer from \method. Similar to \cem, \hdm also suffers a decline in precision, recall, and F1 score in all the cases. Mann-whitney U-test results demonstrate these decreases' significance ($p-value<0.05$) except for precision on the gas pipeline dataset. The effect sizes are strong with a minimum of 0.7211 (recall score on PHM 2015 dataset).
    \item \textbf{DTM effectiveness.} \cemhdm removes the CEM and HDM difficulty measurers, which are computed with DTM information. The comparison between \method and \cemhdm shows the effectiveness of introducing DTM in CL. We can observe from Figure \ref{fig:ablation} that \cemhdm further decreases precision, recall and F1 score in all the cases. All the p-values are smaller than 0.05 except for recall on WADI datasets, while the corresponding effect sizes are strong ($minimum=0.7389$).
\end{itemize}
}


\begin{Summary}{RQ2}{}
\textcolor{black}{We conclude that CL is effective in improving the effectiveness of DTC, and each difficulty measurer (i.e., PDM, CEM, or HDM) contributes to this improvement.}

\end{Summary}

\begin{table*}[h]
\centering
\caption{\textcolor{black}{Statistical test results of the ablation study} }\label{tab:testing_rq2}
\begin{tabular}{|c|c|c|c|c|c|c|c|}

\hline
Datasets & Metrics& Testing & -PDM & -CEM & -HDM & -CEM-HDM & -ALL\\ 
\hline
\multirow{6}{*}{SWaT}  
& \multirow{2}{*}{Precision}  & \cellcolor{Gray} p-value & \cellcolor{Gray}0.0128 & \cellcolor{Gray} <1e-4 & \cellcolor{Gray} <1e-4 & \cellcolor{Gray}2e-04 & \cellcolor{Gray} <1e-4 \\
& & A12 & 0.6811 & 0.78 & 0.9344 & 0.7733 & 1 \\\cline{2-2}
& \multirow{2}{*}{Recall}  & \cellcolor{Gray} p-value & \cellcolor{Gray} <1e-4 & \cellcolor{Gray} <1e-4 & \cellcolor{Gray} <1e-4 & \cellcolor{Gray} <1e-4 & \cellcolor{Gray} <1e-4 \\
& & A12 & 0.8622 & 0.87 & 0.8544 & 0.7922 & 1 \\\cline{2-2}
& \multirow{2}{*}{F1} & \cellcolor{Gray} p-value & \cellcolor{Gray} <1e-4 & \cellcolor{Gray} <1e-4 & \cellcolor{Gray} <1e-4 & \cellcolor{Gray} <1e-4 & \cellcolor{Gray} <1e-4 \\
& & A12 & 0.81 & 0.8778 & 0.94 & 0.85 & 1 \\\hline
\multirow{6}{*}{WADI} 
& \multirow{2}{*}{Precision}  & \cellcolor{Gray} p-value & \cellcolor{Gray}0.0405 & \cellcolor{Gray} <1e-4 & \cellcolor{Gray} <1e-4 & \cellcolor{Gray} <1e-4 & \cellcolor{Gray} <1e-4 \\
& & A12 & 0.3578 & 0.94 & 0.8589 & 1 & 1 \\\cline{2-2}
& \multirow{2}{*}{Recall}  & \cellcolor{Gray} p-value & \cellcolor{Gray}0.2449 & \cellcolor{Gray}0.0024 & \cellcolor{Gray}0.1347 & \cellcolor{Gray}0.0024 & \cellcolor{Gray}0.4161 \\
& & A12 & 0.5611 & 0.74 & 0.58 & 0.7389 & 0.4444 \\\cline{2-2}  
& \multirow{2}{*}{F1} & \cellcolor{Gray} p-value & \cellcolor{Gray}0.0667 & \cellcolor{Gray} <1e-4 & \cellcolor{Gray} <1e-4 & \cellcolor{Gray} <1e-4 & \cellcolor{Gray} <1e-4 \\
 & & A12 & 0.3722 & 0.9433 & 0.8711 & 1 & 1 \\\hline
\multirow{6}{*}{BATADAL} 
& \multirow{2}{*}{Precision}  & \cellcolor{Gray} p-value & \cellcolor{Gray} <1e-4 & \cellcolor{Gray} <1e-4 & \cellcolor{Gray}0.0054 & \cellcolor{Gray} <1e-4 & \cellcolor{Gray} <1e-4 \\
& & A12 & 0.9444 & 0.9844 & 0.6944 & 1 & 1 \\\cline{2-2}
& \multirow{2}{*}{Recall}  & \cellcolor{Gray} p-value & \cellcolor{Gray} <1e-4 & \cellcolor{Gray} <1e-4 & \cellcolor{Gray} <1e-4 & \cellcolor{Gray} <1e-4 & \cellcolor{Gray} <1e-4 \\
& & A12 & 0.9522 & 0.8711 & 0.9167 & 0.8911 & 1 \\\cline{2-2}
& \multirow{2}{*}{F1} & \cellcolor{Gray} p-value & \cellcolor{Gray} <1e-4 & \cellcolor{Gray} <1e-4 & \cellcolor{Gray} <1e-4 & \cellcolor{Gray} <1e-4 & \cellcolor{Gray} <1e-4 \\
& & A12 & 0.9911 & 0.9978 & 0.87 & 1 & 1 \\\hline
\multirow{6}{*}{PHM 2015} 
& \multirow{2}{*}{Precision}  & \cellcolor{Gray} p-value & \cellcolor{Gray} <1e-4 & \cellcolor{Gray} <1e-4 & \cellcolor{Gray} <1e-4 & \cellcolor{Gray} <1e-4 & \cellcolor{Gray} <1e-4 \\
& & A12 & 1 & 1 & 0.9722 & 0.9989 & 1 \\\cline{2-2}
& \multirow{2}{*}{Recall}  & \cellcolor{Gray} p-value & \cellcolor{Gray}0.1142 & \cellcolor{Gray} <1e-4 & \cellcolor{Gray}0.0017 & \cellcolor{Gray} <1e-4 & \cellcolor{Gray} <1e-4 \\
& & A12 & 0.5756 & 0.7567 & 0.7211 & 0.9178 & 0.9978 \\\cline{2-2}
& \multirow{2}{*}{F1} & \cellcolor{Gray} p-value & \cellcolor{Gray} <1e-4 & \cellcolor{Gray} <1e-4 & \cellcolor{Gray} <1e-4 & \cellcolor{Gray} <1e-4 & \cellcolor{Gray} <1e-4 \\
& & A12 & 0.9944 & 0.9922 & 0.9689 & 0.9989 & 1 \\\hline
\multirow{6}{*}{Gas Pipeline} 
& \multirow{2}{*}{Precision}  & \cellcolor{Gray} p-value & \cellcolor{Gray}0.0473 & \cellcolor{Gray}0.685 & \cellcolor{Gray}0.6702 & \cellcolor{Gray}0.3184 & \cellcolor{Gray}0.0011 \\
& & A12 & 0.3311 & 0.4444 & 0.4856 & 0.57 & 0.6944 \\\cline{2-2}
& \multirow{2}{*}{Recall}  & \cellcolor{Gray} p-value & \cellcolor{Gray}0.0277 & \cellcolor{Gray} <1e-4 & \cellcolor{Gray} <1e-4 & \cellcolor{Gray} <1e-4 & \cellcolor{Gray} <1e-4 \\
& & A12 & 0.7122 & 0.9178 & 0.8389 & 1 & 0.9633 \\\cline{2-2}
& \multirow{2}{*}{F1} & \cellcolor{Gray} p-value & \cellcolor{Gray}0.9193 & \cellcolor{Gray}0.001 & \cellcolor{Gray} <1e-4 & \cellcolor{Gray} <1e-4 & \cellcolor{Gray} <1e-4 \\
& & A12 & 0.4989 & 0.7189 & 0.7456 & 0.9456 & 0.9722 \\\hline

\end{tabular}
\end{table*} 


 

\subsection{Results and Analysis for RQ3}\label{subsec:rq3}
\textcolor{black}{
As we target large-scale real-world CPS, efficiency is crucial for \method. As mentioned in Section \ref{subsec:evaluationmetrics}, we use UTT to evaluate the efficiency of \method. Table \ref{tab:et_results} shows the results of the UTT of each method, along with the complexity of each CPS. We can observe that \method improves ATTAIN by 4.2\% ($\frac{9.513-9.114}{9.513}$) in terms of UTT. LSTM-CUSUM takes the least time among all four methods. However, as reported in RQ1-RQ3, LSTM-CUSUM is the least effective method among all.
Table \ref{tab:et_results} further shows that \method takes less time on the SWaT, WADI, and BATADAL datasets, while it takes more time on the PHM 2015 and Gas pipeline datasets. However, we argue that this increase is small and caused by the anomalies' short duration in these two datasets. Short anomalies present smaller complexity, which can train a simpler model quickly. \method is more complicated, as compared to ATTAIN and MAD-GAN, hence the longer UTT. 
}


\begin{table*}[h]
\centering
\caption{Results of efficiency (measured in UTT) of \method and the baselines}\label{tab:et_results}
\begin{tabular}{|c|c|c|c|c|c|}
\hline
Datasets &Complexity & LSTM-CUSUM & MAD-GAN & ATTAIN & \method\\ 
\hline
SWaT         & 0.436 & 3.050 & 8.628 & 8.817 & 8.514 \\ \hline
WADI         & 0.771 & 4.901 & 14.302 & 15.581 & 12.134 \\ \hline
BATADAL      & 0.472 & 4.544 & 8.280 & 8.22 & 7.989 \\ \hline
PHM2015      & 0.180 & 4.944 & 5.533 & 6.117 & 8.539 \\ \hline
Gas Pipeline & 0.479 & 5.282 & 8.591 & 8.   & 8.401 \\ \hline 
Average      & 0.468 & 4.544 & 9.067 & 9.513 & 9.114 \\ \hline    

\end{tabular} 
\end{table*} 

Similar to RQ1-RQ3, we also collected the execution time of 30 runs. We calculate the UTT of these runs and performed the Mann-Whitney U-test, the results of which are reported in Table~\ref{tab:et_testing}. We can observe that LSTM-CUSUM takes significantly less time than \method ($p-value<\alpha$) on all the five datasets, but we exclude it from this comparison due to its low effectiveness. LATTICE takes significantly less time than MAD-GAN and ATTAIN ($p-value<\alpha$) on all the datasets except for the PHM2015 and Gas Pipeline datasets. In terms of A12, the results between \method and ATTAIN are also strong (with the maximum being 0.303) except on the PHM2015 dataset (1.0). The results between \method and MAD-GAN are strong (with the maximum being 0.341) except for PHM2015 (1.0) and Gas Pipeline (0.432). 

\begin{table*}[h]
\centering
\caption{Statistical test results of Unit Training Time (UTT) of \method and the baselines}\label{tab:et_testing}
\begin{tabular}{|c|c|c|c|c|}
\hline 
Datasets & Testing& LSTM-CUSUM & MAD-GAN & ATTAIN\\ 
\hline 
\multirow{2}{*}{SWaT} 
        & p-value & <1e-6 & 8.705e-2 & <1e-6 \\
        & A12&1.0 & 0.341 & 0.0 \\
\hline
  
\multirow{2}{*}{WADI}
        & p-value & <1e-6 & <1e-6 & <1e-6 \\
        & A12     &  1.0  & 0.0   & 0.0\\
\hline

\multirow{2}{*}{BATADAL}
        & p-value & <1e-6 & 3.239e-6 & 4.968e-5 \\
        & A12     &  1.0  &  0.122 & 0.221\\ 
\hline

\multirow{2}{*}{PHM2015}
        & p-value & <1e-6 & <1e-6 & <1e-6\\
        & A12     & 1.0   & 1.0   & 1.0 \\
\hline
 
\multirow{2}{*}{Gas Pipeline}
        & p-value & <1e-6 & \textbf{0.3492} & 7.296e-4 \\
        & A12     & 1.0   & \textbf{0.432 }  & 0.303 \\
\hline

\hline
\end{tabular}
\end{table*}

\begin{Summary}{RQ3}{}
\method generally reduces UTT compared to the baselines. Therefore, we conclude that \method is on par with the baselines in terms of efficiency. \end{Summary}
 
\subsection{Results and Analysis for RQ4}\label{subsec:rq4}
\textcolor{black}{
RQ4 aims to evaluate how early \method can detect an anomaly. We use DDT as the evaluation metric (Section~\ref{subsubsec:metric.ddt}). Table \ref{tab:detection_time} shows the detection time for \method and the baselines on the SWaT, WADI, and BATADAL datasets. We do not include the PHM 2015 and Gas Pipeline datasets in this RQ because their average anomaly lengths are quite small (2 and 4, respectively). DDT degrades to the coarse-grained effectiveness metric for short anomaly detection since a large DDT is equivalent to missing an anomaly.
}

\textcolor{black}{
We present Mann-whitney U-test results and effect sizes in Table~\ref{tab:et_testing}.  
Compared to LSTM-CUSUM, \method's DDT decreases are all significant ($p-value<0.05$) and all effect sizes are strong ($A12=1$). Compared to MAD-GAN, the decreases in WADI and BATADAL are significant ($p-value<0.05$), and effect sizes are strong ($A12=1$ and $A12=0.88$ respectively). However, the SWaT dataset's decrease is not significant ($p-value=0.2621$). Similarly, we observe significant differences for the WADI and BATADAL datasets ($p-value<0.05$) and no significant differences for the SWaT dataset ($p-value=0.3931$) when comparing \method and ATTAIN. The effect sizes on WADI and BATADAL are 0.6578 and 0.9456, respectively, indicating that \method is more likely to detect anomalies faster on these two datasets.
}

\begin{table}[ht] 
    \centering 
    \begin{tabular}{|c|c|c|c|c|}
    \hline
     Dataset &  LSTM-CUSUM & MAD-GAN & ATTAIN & \method\\ \hline
SWaT & 0.2\% & 0.1\% & 0.1\% & 0.1\% \\\hline
WADI & 6\% & 5.7\% & 5.3\% & 5.1\% \\\hline
BATADAL & 4.7\% & 4.5\% & 4.5\% & 4.1\% \\\hline

    \end{tabular}
    \caption{\textcolor{black}{Results of detection delay time} }
    \label{tab:detection_time}
\end{table}

\begin{table}[ht]
    \centering
    \begin{tabular}{|c|c|c|c|c|}
    \hline
    Dataset &Testing & LSTM-CUSUM & MAD-GAN & ATTAIN \\ \hline
\multirow{2}{*}{SWaT} & p-value & <1e-4 & 0.2621 & 0.3931 \\\cline{2-5}
 & A12 & 1 & 0.6133 & 0.4711 \\\hline
\multirow{2}{*}{WADI} & p-value & <1e-4 & <1e-4 & 0.0185 \\\cline{2-5}
 & A12 & 1 & 1 & 0.6578 \\\hline
\multirow{2}{*}{BATADAL} & p-value & <1e-4 & <1e-4 & <1e-4 \\\cline{2-5}
 & A12 & 1 & 0.88 & 0.9456 \\\hline

    \end{tabular}
    \caption{\textcolor{black}{Statistical testing results of detection delay time} }
    \label{tab:detection_time_testing}
\end{table}
\begin{Summary}{RQ4}{}
\method reduces detection delay on all  three datasets when compared with the baselines. Therefore, we conclude that \method is on par with the baselines in terms of detection delay time.
 \end{Summary}


\section{Overall Discussion}\label{sec:disscuss}
\textcolor{black}{
As discussed in the previous sections, \method benefits from CL and digital twin, which allow it to be more effective in anomaly detection. Below, we discuss, in detail, the plausible reasons for \method achieving the effectiveness improvement. 
}

\begin{figure}[ht]
    \centering
    \includegraphics[width=0.7\columnwidth]{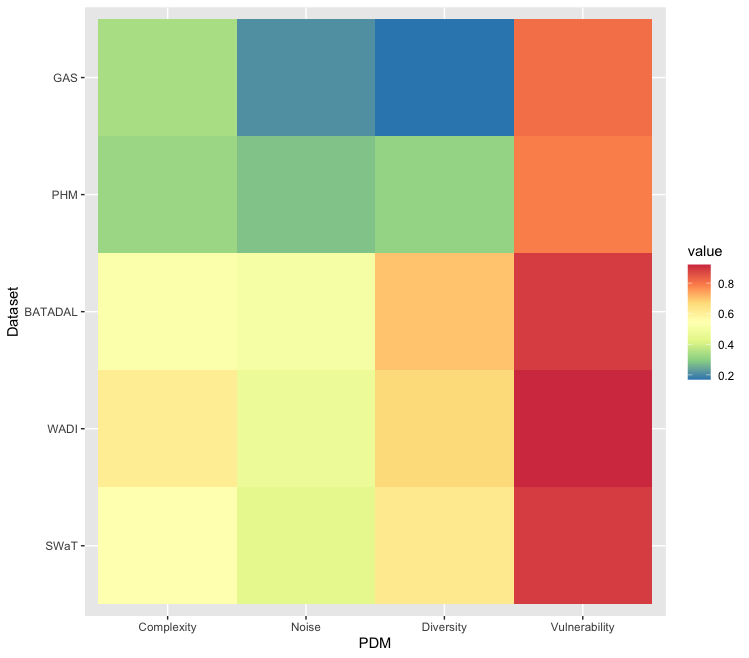}
    \caption{Spearman Correlations between the attack labels and the difficulty measurers}
    \label{fig:corr}
\end{figure}

\noindent\textcolor{black}{
\textbf{Advantage of the predefined difficulty measurers}. In Section \ref{subsubsec:pdm}, we propose a predefined difficulty measurer, consisting of complexity, diversity, noise, and vulnerability. We conducted surveys with these four measurers and found their correlation with the attack labels. We believe capturing this correlation is one of the potential reasons for performance improvement~\cite{kumar2018correlation}.  
We use the Spearman correlation coefficient test for studying the correlation of attack labels with the four measurers. This is a non-parametric test, which is calculated based on the rankings of two variables. The correlation coefficient tends to be high if observations from the two variables have a similar rank position and vice versa. The coefficient value ranges between $-1$ and $1$. Figure \ref{fig:corr} shows the correlation graph between diversity and the attack labels on the SWaT, WADI, BATADAL, and Gas Pipeline datasets. We make a subset containing at least one complete anomaly for each dataset. Correlations are calculated with these subsets instead of the whole datasets for illustrative purpose.
}
\textcolor{black}{
 We find that vulnerability (the last column of Figure~\ref{fig:corr}) has the highest correlation with the attack labels, which is expected because vulnerability is calculated based on the time difference with the attack. Among the other three measurers, diversity has the highest correlation. We find higher correlations in the SWaT, WADI, and BATADAL datasets, while the correlations in the PHM 2015 challenge and Gas pipeline datasets are weaker. We believe this is because the average attack length is much smaller in these two datasets than in the others. Average attack lengths in the PHM 2015 challenge and Gas pipeline datasets are only 2 and 4, respectively. Spearman correlation can hardly capture correlations for such short attacks. However, the high correlations in SWaT, WADI, and BATADAL are sufficient to support our hypothesis that the diversity is related to the attack labels. 
}

\noindent\textbf{Advantage of the automatic difficulty measurers}. We successfully adapted CL to time series data by introducing the contextualized difficulty measurer. Most existing CL approaches use context-free difficulty measurers, implying that the difficulty score of each sample is assigned based on its generic characteristics such as diversity, noise level, and intensity. These context-free difficulty measurers are sufficient for classification tasks in the natural language processing and image recognition domains, while the classification of time series data requires consecutive data for reserving chronological characteristics~\cite{Luo2020,koenecke2019curriculum}. In this paper, we take advantage of DTM and propose two automatic difficulty measurers: CEM and HDM. CEM and HDM successfully incorporate context information into the difficulty scores of each sample. The ablation study in Section \ref{subsec:resultsRQ2} shows the improvement brought by these two contextualized difficulty measurers. Figure \ref{fig:ablation} shows a significant performance drop between \method and \cemhdm.

\noindent\textbf{Advantage of CL's optimization principle.} Another reason for the improvement is the optimization principle of CL. Bengio et al.~\cite{bengiocl} pointed out that CL can be seen as an optimization strategy for non-convex functions. Such a strategy first optimizes a smoother version of the problem to reveal the global picture and then gradually considers less smoothing versions until the target objective of interest is reached. In our case, the introduction of CL potentially prevents our method from getting stuck at a local optimum. As we can observe in the example (Table \ref{tab:running_example}), the first attack starts at 10:29:14 after 29 minutes of normal operation. Consequently, we acquire far more normal data than anomaly data, inducing machine learning methods to a local optimum. CL, however, alleviates this problem by re-ordering batches based on difficulty scores. Particularly, difficult samples are fed into our model gradually, presenting a smoother optimization problem. CL enables \method to learn the global picture instead of being stuck at a local optimum while speeding up the whole training process.

\section{Threats to Validity}\label{sec:threats}
We identify four common types of threats to the validity of our experiments, as discussed below.

\textbf{Conclusion Validity.} We evaluate our method with metrics such as precision, recall, F1, UTT, and DDT. Precision, recall, and F1 are commonly used in classification tasks. However, other metrics, such as ROC and false positive rate, could also be adopted for our evaluation. This could pose a threat to the conclusion validity. These metrics are less commonly used and evaluate the effectiveness of models from similar perspectives as precision, recall, and F1. We will include these metrics in the future if needed. 

Another threat could be that we performed the statistical tests with the Mann-Whitney U-test on samples with a sample size of 30. We are aware that larger sample sizes are always preferred, which however comes with a cost. In the future, when more resources are available, we will conduct experiments with larger sample sizes. 
      
\textbf{Construct Validity.} We also empirically studied whether \method benefits from CL and digital twin for CPS anomaly detection. To that end, we designed \method with digital twin trained with specific CL strategies. However, we are aware that there are other options for digital twin and CL design. Therefore, more experiments are needed to try out various ways of constructing digital twin and different CL strategies, to know better about the potential of CL and digital twin contributing to the performance of \method. 
     
\textbf{Internal Validity.} We compared \method with three baselines: LSTM-CUSUM, MAD-GAN, and ATTAIN, which are the state-of-art in anomaly detection. But we notice that there are other models that can potentially outperform these baselines, such as variational autoencoder and deep belief networks. In the future, we will choose more representative models and conduct more experiments. 
     
\textbf{External Validity.} Our experiments were performed on five testbeds, which are scaled datasets from real operating CPS. Without conducting experiments with real CPS and on different types of CPS, we cannot generalize the performance of \method. However, we would like to point out that, in our context, to conduct experiments with real CPS, \method needs to get connected to them during their operations and obtain their operating data at runtime. Setting up this kind of experiment is complicated and expensive. We plan to work with our collaborators and apply our method in real-world CPS in the future.

 \section{Conclusion and Future Work}\label{sec:conclusion} 
 \method is a novel method, which combines both digital twin and curriculum learning (CL) to address the anomaly detection challenge of CPS. LATTICE extends our previous work ATTAIN. ATTAIN consists of a timed automaton-based digital twin model and a GAN-based digital twin capability. The digital twin model provides ground truth labels indicating whether a CPS is operating in a normal state. Doing so allows ATTAIN to take advantage of a large amount of unlabeled real-time data obtained during the CPS operation and enables ATTAIN to continuously learn along with the operation. We extended ATTAIN by introducing CL to optimize its training process, which forms \method. \textcolor{black}{We performed extensive experiments with \method on five CPS datasets. Experiment results show the performance superiority of \method in comparison to two state-of-art anomaly detectors and ATTAIN, increasing by 0.906\%, 2.363\%, 2.712\%, 2.008\%, 2.367\% on the SWAT, WADI, BATADAL, PHM Challenge 2015, and Gas Pipeline datasets, respectively. We also demonstrated the effectiveness of CL and the different difficulty measurers with the ablation study. Finally, we demonstrated that \method is on par with the baselines in terms of the training time and detection delay time.} 

In the future, we plan to conduct more experiments on real-world CPS of various domains to evaluate the scalability and generalization of \method. We will also consider exploring digital twins for more challenging tasks, such as detecting attacks targeting multiple CPS simultaneously which requires developing an integrated digital twin model. 
\section*{Acknowledgement}\label{sec:ack}
The SWaT, WADI, and BATADAL datasets were provided by iTrust, Centre for Research in Cyber Security, Singapore University of Technology and Design. The research presented in this paper has benefited from the Experimental Infrastructure for Exploration of Exascale Computing (eX3), which is financially supported by the Research Council of Norway under contract 270053. Qinghua Xu is supported by a project funded by the Norwegian Ministry of Education and Research. Shaukat Ali and Tao Yue are supported by Horizon 2020 project ADEPTNESS (871319) funded by the European Commission. 
\bibliographystyle{tosem}
\bibliography{tosemDT.bib}


\end{document}